\documentclass[submission,copyright,creativecommons]{eptcs}
\providecommand{\event}{CAPNS 2018} 
\usepackage{bbm}
\usepackage{amsmath,amssymb, amsthm}
\usepackage[abs]{overpic}
\usepackage{caption}
\usepackage{float}
\usepackage{subcaption}
\usepackage{mathrsfs}
\usepackage{soul}
\usepackage{afterpage}
\usepackage{array}
\usepackage{tikz-cd}
\usepackage{longtable}
\usepackage[toc,page]{appendix}
\usetikzlibrary{shapes}
\usepackage{tikz}
\usepackage{tikz-qtree}
\usepackage{forest}
\usepackage{mathtools}
\usetikzlibrary{decorations.pathmorphing}
\usepackage{pifont}
\usepackage{mdframed}
\usepackage{wasysym}
\usepackage{graphicx}

\usepackage{etoolbox,xspace,bbm}
\usepackage{hyperref}

\pgfdeclarelayer{nodelayer}
\pgfdeclarelayer{edgelayer}
\pgfsetlayers{nodelayer,edgelayer,main}

\allowdisplaybreaks

\newmdenv[
  innermargin=0pt,
  outermargin=0pt,
  innerleftmargin=0pt,
  innerrightmargin=0pt,
  innertopmargin=0pt,
  innerbottommargin=0pt,
  topline=false,
  rightline=false,
  leftline=false,
  bottomline=false,
   ]{tipbox*}

\theoremstyle{plain}
\newtheorem{theorem}{Theorem}[section]

\newtheorem{corpus}[theorem]{Corpus}
\newtheorem{remark}[theorem]{Remark}
\newtheorem{definition}[theorem]{Definition}

\theoremstyle{remark}
\newtheorem{example}[theorem]{Example}

\AtEndEnvironment{remark}{\hfill$\lozenge$\\}
\AtEndEnvironment{const}{\hfill$\lozenge$}
\AtEndEnvironment{nonexample}{\hfill$\lozenge$}
\AtEndEnvironment{corpus}{\hfill$\square$}

  \forestset{
  colour my roots/.style={
    before typesetting nodes={
      edge=#1,
      for ancestors={
        edge=#1,
        #1,
      },
      #1,
    }
  },
  my edge label/.style={
    edge label={
      node [midway, fill=white, font=\scriptsize] {#1}
    }
  }
}

\newcommand{\defeq}{\mathrel{\mathop:}=}

\definecolor{grn}{RGB}{0, 160, 0}

\title{Applying Distributional Compositional Categorical Models of Meaning to Language Translation}
\author{Brian Tyrrell
\institute{Department of Mathematics\\
University of Oxford}
\email{brian.tyrrell@maths.ox.ac.uk}
}

\def\titlerunning{Appyling DisCoCats to Language Translation}
\def\authorrunning{B.\ Tyrrell}
\begin{document}
\maketitle
\forestset{qtree edges/.style={for tree={
parent anchor=south, child anchor=north}}}

\begin{abstract}
The aim of this paper is twofold: first we will use vector space distributional compositional categorical models of meaning to compare the meaning of sentences in Irish and in English (and thus ascertain when a sentence is the translation of another sentence) using the cosine similarity score. Then we shall outline a procedure which translates nouns by understanding their context, using a conceptual space model of cognition. We shall use metrics on the category \textbf{ConvexRel} to determine the distance between concepts (and determine when a noun is the translation of another noun). This paper will focus on applications to Irish, a member of the Gaelic family of languages.
\end{abstract}

\section{Introduction}
The raison d'\^{e}tre of \underline{Dis}tributional \underline{Co}mpositional \underline{Cat}egorical (henceforth referred to as \underline{DisCoCat}) Models of Meaning originates in the oft quoted mantra of the field:

\vspace{2mm}
\begin{tabular}{|p{0.9\textwidth}}
\textit{``You shall know a word by the company it keeps.''}\\
-John R. Firth, \textit{A synopsis of linguistic theory 1930-1955}, (1957).
\end{tabular}
\vspace{2mm}

The broad idea of such models in natural language processing is to marry the semantic information of words with the syntactic structure of a sentence using category theory to produce the whole meaning of the sentence. The semantic information of a word is captured (in early models, cf.\ \cite{bob, greffen3, greffen2}) by a vector in a tensor product of vector spaces using a corpus of text to represent a given word in terms of a fixed basis of other words; i.e.\ by distributing the meaning of the word across the corpus. In later models \cite{dan} convex spaces are used instead of vector spaces in an effort to better capture the representation of words in the human mind. 

It is the focus of this paper to exploit the existing DisCoCat structure for language translation. We shall use a vector space model of meaning, defined by Coecke et al.\ \cite{bob} and introduced in \emph{Section \ref{vect}}, to assign meaning to sentences in English and then in Irish. These meanings are then compared via an inner product on the shared sentence space of English and Irish vector space models of meaning in \emph{Section \ref{irishvect}}. We discuss the results of this on example sentences in \emph{Section \ref{irishvect}} \&\ \emph{Appendix \ref{similarity}}. The next three sections of the paper focus on using conceptual spaces (or \emph{concepts}) in place of vector spaces to understand the meaning of nouns. In \emph{Section \ref{claso}} we work on a system of adjective \&\ noun classification which, in \emph{Section \ref{five}}, leads to the generation of convex spaces representing noun concepts from a given corpus. While other authors (cf.\ Derrac \&\ Schockaert \cite{derrac}) have also induced conceptual spaces algorithmically from corpora, the treatment we propose is tailored towards the authors use of metrics in \emph{Section \ref{7}} to determine the `distance' between concepts coming from different languages. The system we will propose cannot capture every type of adjective, however it is sufficiently complex and complete to allow us to start analysing text in a meaningful way.

Before this, we must determine the Lambek pregroup grammar structure for Irish (which does not exist in the current literature) and, as we shall see in \emph{Section \ref{irishlambek}}, is nontrivial in some aspects. We will only deal with an elementary fragment of Irish, in much the same way an elementary fragment of English is used in \cite{bob}. The ideas presented in this paper can be applied to many other languages, however the author has chosen Irish due to its relative rareness in literature and its high regularity in verb structure. For instance, across all of Irish there exist exactly eleven irregular verbs; with the exception of these eleven, every other verb can be conjugated in an extremely efficient and easy manner.

\bigskip
\section{Lambek Pregroup Grammar Structure for Irish}\label{irishlambek}

Coecke et.\ al.\ use the Lambek pregroup grammar structure to determine maps which assign meanings to sentences \cite[\S 3.5]{bob}. This is all built exclusively through English, but there are no barriers to moving to a different language; Lambek et al.\ \cite{lambekbargelli, lambekbargelli2, lambekcasadio, lambekpreller} detail a pregroup structure for French, Arabic, Latin and German, respectively. However the author cannot find evidence of the same treatment in Irish. Thus, in order to create Irish DisCoCat models, we must create a `Lambek Pregroup Grammar' for the language.

\subsection{Irish Grammatical Structure}\label{ista}
For our purposes, we do not need a structure as complicated as Lambek's work \cite{lambek}, rather we shall mirror the English approach: four basic types - nouns ($n$), declarative statements ($s$), infinitives of verbs ($j$) and gluing types ($\sigma$). We hand construct the following compound types:
\begin{enumerate}
    \item \textbf{Transitive verbs} are assigned the type $s n^l n^l$. This is because Irish follows the rule \emph{Verb Subject Object}. The only exception to this is the copula \textit{is}, which we also assign the type $s n^l n^l$ but note the sentence structure must be \textit{Verb Object Subject}. This verb-like word is used in sentences that state equivalences between, or crucial attributes of, the subject and object. 
    
    For example, even though the Irish for the verb ``to be'' is \textit{b\'{i}}, which in the present tense is \textit{t\'{a}}, one would say
    \begin{align*}
    &\mbox{ \textit{Is docht\'{u}ir m\'{e}}} \quad &&\mbox{for} \quad \mbox{``I am a doctor''} \quad \mbox{and}\\
    &\mbox{ \textit{T\'{a} scamaill sa sp\'{e}ir}} \quad &&\mbox{for} \quad \mbox{``There are clouds in the sky'',}
    \end{align*}
    
    \item \textbf{Adjectives} are assigned the type $n^r n$. This is because Irish follows the rule \emph{Noun Adjective}.    
    
    \item \textbf{Adverbs} are assigned the type $s^r s$; they appear at the end of sentences.
    
    \item \textbf{Prepositions} as \textit{whole phrases} are assigned the type $n^r n$. This is because Irish follows the rule \emph{Preposition Noun}, as in English, so we give the same type assignment as in \cite{greffen3}. Note that prepositions in Irish always come before the noun, and adjectives after, so we cannot confuse them.
    
    It should be noted that Irish (sometimes) modifies the noun after a preposition directly by inserting an \textit{ur\'{u}} or \textit{s\'{e}imhi\'{u}} into the noun - additional letters to change the pronunciation of the word. So, for example, whilst \textit{table} in Irish is \textit{bord}, \textit{on the table} becomes \textit{ar an \textbf{m}bord}. This is a sign that correlates with the change in type assignment of the affected noun.
\end{enumerate}
Consider the following demonstration:

\vspace{2mm}
\begin{example}
English sentence: ``I broke the vase under the large bridge yesterday'' In Irish: ``\textit{Bhris m\'{e} an v\'{a}sa faoin droichead m\'{o}r inn\'{e}}''. This has the type assignment
    \begin{gather*}
    \mbox{Bhris} \quad \mbox{m\'{e}} \quad \mbox{an v\'{a}sa} \quad \mbox{faoin droichead}\quad \mbox{m\'{o}r} \quad \mbox{inn\'{e}} \\    
    sn^l n^l \qquad n \qquad n \qquad n^r n \qquad n^r n \qquad s^r s\tag*{$\lozenge$}
    \end{gather*}
\end{example}

\vspace{3mm}
Finally, Sadrzadeh et al.\ \cite{sadrz} consider subject relative pronouns (such as \emph{who(m)} and \emph{which}) and object relative pronouns (such as \emph{that}). They assign the pregroup types as follows:
$$n^r n s^l n \mbox{   (subject relative pronoun)} \qquad n^r n n^{ll} s^l \mbox{  (object relative pronoun)}$$

However, in Irish these particular words (\emph{who(m), which, that}) are simply represented by one word: {\textit{a}}. Moreover, the grammatical structure of a sentence containing these relative pronouns is the same regardless of whether the relative pronouns are object or subject modifying; the sentence structure is always ``noun relative-pronoun verb noun''. So for Irish we can define:
\begin{itemize}
    \item[(5)] \textbf{Relative Pronouns.} Let $n^r n n^{ll} s^l$ be the pregroup type of {\emph{a}}, the Irish relative pronoun \emph{who(m), which,} and \emph{that}. 
\end{itemize}
This concludes the work required to use a pregroup grammar structure in Irish.

\bigskip
\section{A Vector Space based Model of Meaning}\label{vect}
We will now create a vector space model of meaning from \emph{Corpus \ref{corpa1}}, located in \emph{Appendix \ref{sweng}}. The section after this will create another vector space model of meaning, this time in Irish, from the translation of \emph{Corpus \ref{corpa1}}. The underlying principal is that once we have the meaning of a sentence in an abstract vector space $S$, it does not matter what the language of the sentence is, as it can be compared via an inner product on $S$. An application of this idea is to measure the accuracy of translation tools such as \emph{Google Translate}, and also to potentially train software (off large corpora) to accept input commands in any language. 

The corpus of text chosen by the author is a modified copy of the plot of \emph{Star Wars: Episode III - Revenge of the Sith} obtained from Wikipedia. The full corpus of text is presented in \emph{Appendix \ref{sweng}}. We shall closely follow the exposition presented by Grefenstette and Sadrzadeh \cite{greffen, greffen2} throughout.

As we are primarily interested in the vector space $N$ of nouns, we shall begin there. We define the basis to consist of the five most commonly occurring words against which we shall measure all other nouns in the corpus:
$$\mbox{Basis of $N$}=\{\mbox{Anakin, Palpatine, Jedi, Obi-Wan, arg-evil}\},$$
where `arg-evil' denotes the argument of the adjective `evil' (cf.\ \cite[\S 3]{greffen3}). The coordinates of a noun $K$ follow from counting the number of times each basis word has appeared in an $m$ word window around $K$; in particular, $K$ is given a coordinate of $k$ for `arg-evil' if $K$ has appeared within $m$ words of a noun described as `evil' in the same sentence, $k$ times in the corpus. For this paper, set $m=3$. In this basis

\vspace{-3mm}
\begin{equation*}
\begin{aligned}[c]
&\mbox{Anakin} = [1, 0, 0, 0, 0],\\
    &\mbox{Palpatine} = [0, 1, 0, 0, 0],\\
    &\mbox{Jedi} = [0,0,1,0,0],\\
    &\mbox{Obi-Wan}  = [0, 0, 0, 1, 0],\\
    &\mbox{arg-evil} = [0,0,0,0,1],
\end{aligned}
\qquad
\begin{aligned}[c] 
    &\mbox{Padm\'{e}} = [5, 0, 0, 2, 0],\\
    &\mbox{Yoda} = [0, 2, 1, 3, 1],\\
    &\mbox{Emperor} = [1, 5, 0, 1, 1],\\
    &\mbox{mastermind} = [2, 2, 0, 0, 1],\\   
    &\mbox{evil (noun)} = [0,2,1,0,0]\\
\end{aligned}
\qquad
\begin{aligned}[c] 
    &\mbox{Mace Windu} = [1, 1, 2, 0, 0],\\
    &\mbox{Sith Lord} = [1, 1, 0, 0, 1],\\
    &\mbox{General Grievous} = [0, 1, 4, 1, 1],\\
    &\mbox{dark side of the Force} = [4, 2, 1, 1, 1],\\
    &\mbox{ }\\
\end{aligned}
\end{equation*}
\noindent where we treat `dark side of the Force', as one noun. It has appeared within 3 words of `Anakin' 4 times, `Palpatine' 2 times, `Jedi' once, `Obi-Wan' once, and the argument of `evil' once (see \emph{Corpus \ref{corpa1}}).

As described by Grefenstette and Sadrzadeh \cite{greffen} there exists an exact procedure for learning the weights for matrices of words. If we assume $S = N \otimes N$ (so the basis of $S$ is of the form $\overrightarrow{n}_i \otimes \overrightarrow{n}_k$) then the meaning vector of a transitive verb in a sentence is
$$\overrightarrow{\mbox{subject  verb  object}} = \overrightarrow{verb} \odot (\overrightarrow{subject} \otimes \overrightarrow{object}),$$
where $\odot$ is the Kronecker product. A transitive verb is described by a two dimensional matrix; using the data of \emph{Corpus \ref{corpa1}},

\[
C^{\mbox{\tiny  turn}} = 
\begin{bmatrix}
    8 & 7 & 4 & 2  & 2 \\
    1 & 0 & 0 & 0  & 0 \\
    4 & 2 & 1 & 1  & 1 \\
    0 & 2 & 1 & 3  & 1 \\
    0 & 0 & 0 & 0  & 0
\end{bmatrix},
\qquad C^{\mbox{\tiny  is}} = 
\begin{bmatrix}
    0 & 0 & 1 & 0  & 1 \\
    5 & 4 & 1 & 1  & 3 \\
    4 & 0 & 1 & 4  & 0 \\
    1 & 0 & 3 & 1  & 0 \\
    1 & 0 & 1 & 1  & 1
\end{bmatrix}.
\]
\vspace{0.5mm}

We only use sentences from \emph{Corpus \ref{corpa1}} which have a transitive use of ``is'', e.g.\ ``Anakin is a powerful Jedi'' as opposed to ``he is too powerful'' in our calculation of $C^{\mbox{\tiny  is}}$. An adjective $A$ can be determined in the same fashion, so an adjective vector is computed to be the sum of the vectors of its arguments, e.g.\ $C^{\mbox{\tiny powerful}} = [1,2,2,1,1]$, or $C^{\mbox{\tiny brave}} = [7, 1, 2, 5, 0]$. 

\subsection{Representing a sentence as a vector.}\label{warmup}

Now consider the sentence at the start of \emph{Corpus \ref{corpa1}}:
$$\mbox{\textbf{Palpatine is a mastermind who turns Anakin to the dark side of the Force}.}$$
Let us calculate a meaning vector for this sentence. The prepositional phrase ``to the dark side of the Force'' (abbreviated as ``to DSOF'') is represented as a vector and is given by the sum of the vectors of its arguments; $C^{\mbox{\tiny  to DSOF}} = [3, 0, 1, 0, 0]$. The sentence ``\emph{Palpatine is a mastermind who turns Anakin to the dark side of the Force}'' has the type assignment $\qquad n \quad  n^r s n^l \quad n \quad n^r n s^l n \quad n^r s n^l \quad n \quad n^r n$, $\quad$ using the convention from \cite{greffen} that the prepositional phrase ``to the dark side of the Force'' as a whole has the assignment $n^r n$. The sentence has a meaning vector of
\vspace{-5mm}
\begin{align*}
&\overrightarrow{\mbox{Palpatine is a mastermind who turns Anakin to the dark side of the Force}}\\
&= \sum_{k,p,r,v,x} c_{k}^{\mbox{\tiny  Palp}} c_{kpr}^{\mbox{\tiny  is}} c_r^{\mbox{\tiny  mm}} c_{rx}^{\mbox{\tiny  turns}} c_{v}^{\mbox{\tiny  Anakin}} c_{vx}^{\mbox{\tiny  to DSOF}} \overrightarrow{s}_p \quad= \quad\sum_p (64c_{2p1}^{\mbox{\tiny  is}} + 8c_{2p2}^{\mbox{\tiny  is}})\overrightarrow{s}_p. \tag{\ding{168}}
\end{align*}

This does not have much meaning, as $S$ is an arbitrary vector space. 
If we set $S = N \otimes N$, then $\overrightarrow{s}_p = \overrightarrow{n}_i \otimes \overrightarrow{n}_j$. The verb matrix $C^{\mbox{\tiny  is}}$ is given above, thus $(\mbox{\ding{168}}) = 320 \overrightarrow{n}_2 \otimes \overrightarrow{n}_1 + 32\overrightarrow{n}_2 \otimes \overrightarrow{n}_2$. This sum of tensor products only becomes meaningful when we are comparing sentences via an inner product on $S$, as we do next.

Consider the sentence ``\emph{Mace Windu is a mastermind who turns Anakin to the dark side of the Force}''. When we compare this sentence to ``\emph{Palpatine is a mastermind who turns Anakin to the dark side of the Force}'' we obtain a similarity score of $\tfrac{103424}{\sqrt{41066496\cdot 103424}} = 0.53$. This is calculated by taking the inner product of the two sentences (103424) and dividing by the square root of the product of their lengths.

However, if we compare ``\textit{The Emperor} is a mastermind who turns Anakin to the dark side of the Force'' to ``Palpatine is a mastermind who turns Anakin to the dark side of the Force'' we obtain a much higher similarity score of 0.99. Of course, as Palpatine is the Emperor, the similarity should be very high! A similar inner product with the sentence ``\textit{Padm\'{e}} is a mastermind who turns Anakin to the dark side of the Force'' gives a similarity score of 0; as expected these sentences are not similar at all, as ``Padm\'{e}'' is very different to ``Palpatine''. 

The vector space model of meaning has managed to extract these key themes from the corpus. Now our goal is to extract the same key ideas from an Irish corpus.

\bigskip
\section{Bilingual Sentence Comparison via the Vector Space Model of Meaning}\label{irishvect}

We shall now compare sentences between corpora in different languages. Our Irish vector space model of meaning shall be created from \emph{Corpus \ref{corpb1}}, using the methods detailed in the previous section.

The calculations in \emph{Section \ref{vect}} \&\ \emph{Appendix \ref{similarity}} require $S = N \otimes N$; to that end let the basis of $N'$, the Irish noun space, be $\{\mbox{\it Anakin, Palpatine, Jedi, Obi-Wan, arg-olc}\}$, where ``\textit{arg-olc}'' corresponds to the argument for the adjective \textit{olc} - in English, `evil'. This is also the collection of the five most commonly occurring nouns in \emph{Corpus \ref{corpb1}} exactly (which might not really be a surprise as \emph{Corpus \ref{corpb1}} is a translation of \emph{Corpus \ref{corpa1}}, and nouns in English typically have one translation to Irish). 

Take for example the sentence ``\textit{Palpatine is an evil Emperor}''. In Irish, this is ``\textit{Is Impire olc \'{e} Palpatine}''. The sentence has the type assignment and reduction diagram

\vspace{-3mm}
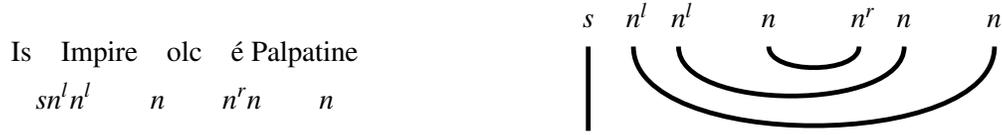
\begin{minipage}{0.4\textwidth}
\begin{gather*}
    \mbox{Is} \quad \mbox{Impire} \quad \mbox{olc} \quad \mbox{\'{e} Palpatine} \\    
    sn^l n^l \qquad n \qquad  n^r n \qquad  n 
\end{gather*}
\end{minipage}
\begin{minipage}{0.6\textwidth}
\begin{figure}[H] \centering
\begin{tikzpicture}[scale=0.6]
	\begin{pgfonlayer}{nodelayer}
		\node  (0) at (-8, 1) {${}^{\phantom{l}}_{\phantom{l}}s{}^{\phantom{l}}$};
		\node  (1) at (-7, 1) {${}^{\phantom{l}}_{\phantom{l}}n^l$};
		\node  (2) at (-6, 1) {${}^{\phantom{l}}_{\phantom{l}}n^l$};
		\node  (3) at (-4, 1) {${}^{\phantom{l}}_{\phantom{l}}n{}^{\phantom{l}}$};
		\node  (4) at (-2, 1) {${}^{\phantom{l}}_{\phantom{l}}n^r$};
		\node  (5) at (-1, 1) {${}^{\phantom{l}}_{\phantom{l}}n{}^{\phantom{l}}$};
		\node  (6) at (1, 1) {${}^{\phantom{l}}_{\phantom{l}}n{}^{\phantom{l}}$};
		\node  (7) at (-8, -1.7) {};
	\end{pgfonlayer}
	\begin{pgfonlayer}{edgelayer}
		\draw [bend right=90, looseness=0.8, line width = 1.8pt] (3) to (4);
		\draw [line width = 1.8pt](0) to (7);
		\draw [bend right=90, looseness=0.75, line width = 1.8pt] (2) to (5);
		\draw [bend right=90, looseness=0.75, line width = 1.8pt] (1) to (6);
	\end{pgfonlayer}
\end{tikzpicture}
\end{figure}
\end{minipage}

corresponding to a map
$$f = (1_S \otimes \epsilon_N) \circ (1_S \otimes 1_N \otimes \epsilon_N \otimes 1_N) \circ (1_S \otimes 1_N \otimes 1_N \otimes \epsilon_N \otimes 1_N \otimes 1_N).$$

Therefore the sentence ``\textit{Is Impire olc \'{e} Palpatine}'' is assigned the following meaning vector:
$$\overrightarrow{\mbox{Is Impire olc \'{e} Palpatine}} = \sum_{ijk, p} c_{ijk}^{\mbox{\tiny   Is}} c_p^{\mbox{\tiny  Imp}} c_{pk}^{\mbox{\tiny  olc}} c_j^{\mbox{\tiny  Palp}} \overrightarrow{s}_i.$$

In order to evaluate this sentence, we need values for $c_{ijk}^{\mbox{\tiny   Is}}, c_p^{\mbox{\tiny  Imp}}, c_{pk}^{\mbox{\tiny  olc}}$, and $c_j^{\mbox{\tiny  Palp}}$. The adjectives and adverbs are calculated in the same way as in \S\ref{vect}. We must be a little more careful with verbs. The copula \emph{is} follows the rule \emph{Is Object Subject}, and all other transitive verbs follow the rule \emph{Verb Subject Object}, so we define
$$\overrightarrow{is} \defeq \sum_{ij} \overrightarrow{subject}_i \otimes \overrightarrow{object}_j, \qquad \overrightarrow{verb} \defeq \sum_{ij} \overrightarrow{object}_i \otimes \overrightarrow{subject}_j,$$
and also that
\vspace{-3mm}
\begin{gather*}
    \overrightarrow{\mbox{is object subject}} \defeq \overrightarrow{is} \odot (\overrightarrow{subject} \otimes \overrightarrow{object}),\\
    \overrightarrow{\mbox{verb subject object}} \defeq \overrightarrow{verb} \odot (\overrightarrow{object} \otimes \overrightarrow{subject}).
\end{gather*}

Note the copula is treated the same as in English\footnote{One small caveat: ``\textit{is}'' is translated to have the same meaning as the verb ``to be'' in English, which in Irish corresponds to the verb ``\textit{b\'{i}}'', which in \emph{Corpus \ref{corpb1}} is conjugated as ``\textit{t\'{a}}''. The result? $C^{\mbox{\tiny  Is}} = C^{\mbox{\tiny  t\'{a}}}$ is calculated by including sentences with use of either ``\textit{T\'{a}}\dots'' or ``\textit{Is}\dots''.}. Now, let us compare sentences between Irish and English. We obtain the following data from \emph{Corpus \ref{corpb1}}. (Note ``\emph{Impire}'' is ``Emperor'', ``\emph{Tiarna Sith}'' is ``Sith Lord'', ``\textit{taobh dorcha na F\'{o}rsa}'' is ``dark side of the Force'', and ``\textit{cr\'{o}ga}'' is ``brave''.)

\vspace{-2mm}
\begin{minipage}{0.33\textwidth}
\begin{align*}
&\mbox{Impire} = [1,5,0,1,1],\\
&\mbox{Palpatine} = [0,1,0,0,0]\\
&\mbox{{taobh dorcha na F\'{o}rsa}} = [5,4,1,1,1],\\
&C^{\mbox{\tiny  olc}} = [2, 5, 2, 1, 3],
\end{align*}
\end{minipage}
\begin{minipage}{0.3\textwidth}
\begin{align*}
&\mbox{Yoda} = [0,1,2,3,0],\\
&\mbox{Tiarna Sith} = [1,0,0,0,2],\\
&\mbox{Padm\'{e}} = [5,0,0,1,0],\\
&C^{\mbox{\tiny  cr\'{o}ga}} = [7,1,2,4,0].
\end{align*}
\end{minipage}
\begin{minipage}{0.36\textwidth}
\begin{align*}
C^{\mbox{\tiny  Is}} = 
\begin{bmatrix}
    6 & 0 & 1 & 0  & 1 \\
    6 & 4 & 1 & 0  & 2 \\
    2 & 0 & 2 & 0  & 2 \\
    0 & 0 & 3 & 0  & 0 \\
    3 & 0 & 0 & 0  & 2\\
\end{bmatrix}. \hspace{10mm}  
\end{align*}
\end{minipage}
\vspace{2mm}

It is quite welcome that $C^{\mbox{\tiny  Is}}$ is different to the English $C^{\mbox{\tiny  is}}$, as in Irish the verb ``to be'' is sometimes used in conjunction with another verb, which becomes the main transitive verb of the sentence. Thus, there are fewer occurrences of ``\textit{t\'{a}}'' or ``\textit{is}'' in \emph{Corpus \ref{corpb1}} than ``is'' in \emph{Corpus \ref{corpa1}}.

The result of our two assumptions (that $S = N \otimes N$ and the basis of $N'$ is the exact translation of the basis of $N$) is we can meaningfully compare the following sentences. In the first instance,
\begin{align*}
&\langle \mbox{Palpatine is an evil Emperor } | \mbox{ \it Is Impire olc \'{e} Palpatine}\rangle = 10174.
\end{align*}

The length of the former is 10182 and the length of the latter is 10180, meaning the similarity score between the sentence ``Palpatine is an evil Emperor'' and its Irish translation
``\textit{Is Impire olc \'{e} Palpatine}'' is $\tfrac{10174}{\sqrt{10182 \cdot 10180}} = 0.99$; very high. On the other hand, if we try to compare sentences that are not translates of one another, say ``Yoda is a powerful Jedi'' to ``\textit{Is Jedi cr\'{o}ga \'{e} Palpatine}'' (in English, ``Palpatine is a brave Jedi''), we receive low scores\footnote{Technically when calculating $C^{\mbox{\tiny  cr\'{o}ga}}_{ij}$ - the matrix for the adjective `brave' - we are also counting the various different translations of ``brave'' occurring in \emph{Corpus \ref{corpb1}}, such as ``{\it go crua}'' or ``{\it go l\'{a}idir}''.}:
\begin{align*}
&\langle \mbox{Yoda is a powerful Jedi } | \mbox{ \it  Is Jedi cr\'{o}ga \'{e} Palpatine}\rangle = 8.
\end{align*}

The length of the former is 348 and the length of the latter is 4, so the similarity score between the sentence ``Yoda is a powerful Jedi'' and ``\textit{Is Jedi cr\'{o}ga \'{e} Palpatine}'' is $\tfrac{8}{\sqrt{348 \cdot 4}} = 0.21$; relatively low.

The following table houses a collection of sentence pairs and their similarity scores, calculated according to this method. The reader is invited to see \emph{Appendix \ref{similarity}} for a more complicated sentence comparison.

\vspace{-1mm}
\begin{center}
 \begin{tabular}{| c | c | c |} 
 \hline
 English Sentence & Irish Translation & Similarity Score\\ [0.5ex] 
 \hline\hline
 Yoda is a powerful Jedi & \textit{Is Jedi cumhachtach \'{e} Yoda} &  0.94\\
 \hline
 Palpatine is an evil Emperor & \textit{Is Impire olc \'{e} Palpatine} & 0.99 \\
 \hline
 A brave Padm\'{e} turns to Anakin & \textit{Casann Padm\'{e} cr\'{o}ga chuig Anakin} & 1  \\
 \hline
 Obi Wan turns to the powerful Yoda & \textit{Casann Obi-Wan go Yoda cumhachtach} & 0.87 \\
 \hline
 Padm\'{e} is a brave Jedi & \textit{Is Jedi cr\'{o}ga \'{e} Padm\'{e}} & 0.94 \\
 \hline
 Anakin is a Sith Lord & \textit{Is Tiarna Sith \'{e} Anakin} & 0.32 \\ 
 \hline
 The Jedi turn to the brave Mace Windu & \textit{Casann na Jedi go Mace Windu cr\'{o}ga} & 0.99 \\ [0.5ex]
 \hline
\end{tabular}
\end{center}
\newpage

The BLEU score, introduced in \cite{bleu}, provides a measure of how accurate a machine translation system is; in the below table we demonstrate how the BLEU score of two sentences with minor variations compares to the similarity score of the sentences. This score is calculated as follows: let $R$ be a reference sentence and $C$ be a candidate sentence. The goal is to measure how the \emph{n-grams} (blocks of $n$ words in succession) appear and compare in $R$ and $C$. To that end, define
$$p_n = \frac{\sum_{\mbox{\tiny n-gram}\in C}\mbox{ } \min\{\#\mbox{\small times n-gram appears in $C$, }\#\mbox{\small times n-gram appears in $R$}\}}{\sum_{\mbox{\tiny n-gram}\in C}\mbox{ }\#\mbox{\small times n-gram appears in $C$}}.$$
We can also introduce a \emph{brevity penalty} (BP) if the candidate is too short: this is $\exp(1 - \tfrac{|R|}{|C|})$ if $|C| \leq |R|$ and 1 otherwise. Finally, the $\mbox{BLEU score} \defeq \mbox{BP} \cdot \exp\left(\sum_{n=1}^4 \tfrac{1}{4} \log p_n \right)$. If there is no n-gram overlap between $C$ and $R$ for any order of n-grams ($1 \leq n \leq 4$), BLEU returns the value 0. The author used smoothing function 7 of \cite{chencherry} to avoid this harsh behaviour. 

\begin{center}
 \begin{tabular}{| c | c | c | c |} 
 \hline
 Reference Sentence& Candidate Sentence & Similarity Score & BLEU Score\phantom{\hspace{0.25mm}}\\ [0.5ex] 
 \hline\hline
 Yoda is a powerful Jedi & Yoda turns to the powerful Jedi &  0.95 & 0.32\\
 \hline
 Anakin is a Sith Lord & Obi-Wan is a Sith Lord & 0 & 0.7 \\
 \hline
 \textit{Is Impire olc \'{e} Palpatine} & \textit{Is Impire olc \'{e} Mace Windu} & 0.98 & 0.7 \\
 \hline
 \begin{tabular}{@{}c@{}}\textit{Casann na Jedi go Mace Windu}\\ \textit{cumhachtach}\end{tabular} & \begin{tabular}{@{}c@{}}\textit{Casann Ginear\'{a}l Grievous go}\\ \textit{Mace Windu cr\'{o}ga}\end{tabular} & 0.76 & 0.27 \\ [0.5ex]
 \hline
\end{tabular}
\end{center}

\vspace{2.25mm}
In \cite{dan} the authors work with conceptual spaces: instead of nouns being labelled relative to nouns they appear often with, instead nouns are represented by other words that describe them. The hope would be this removes instances of problematic translation, like in the sixth example of the first table. However, building on the ideas of Bolt et al.\ \cite{dan} and G\"{a}rdenfors \cite{gardenforce} much work would need to be done to capture the intricacies presented here. In the next sections we will instead tackle a simpler example involving planets and fruit.

\section{Word Classification}\label{claso}
According to Dixon and Aikhenvald \cite{dixon}, ``three word classes are \dots implicit in the structure of each human language: nouns, verbs and adjectives.'' It is the goal of this section to specify a treatment of nouns and adjectives for use in conceptual space creation. Once we have some sort of classification system for each of these, we can proceed with creating a conceptual space from a given corpus. For example, in the case of adjectives we wish to classify words such as `heavy', `red' or `hot', and to each assign a numeric value that transcends language and thus can be compared across (say) Irish and English.

\subsection{Adjectives}\label{adj}
In their landmark work, Dixon and Aikhenvald \cite{dixon} give a complete treatment of adjective classes as they arise in various languages across the globe, such as Japanese, Korean, Jarawara, Mam and Russian. In particular, they name seven core types of adjectives that consistently and naturally arise:

\begin{equation*}
\begin{aligned}[c]
    &\mbox{(1) \textbf{Dimension.} (big, small, short, tall, etc.)}\\
    &\mbox{(2) \textbf{Age.} (new, old, etc.)}\\
    &\mbox{(3) \textbf{Value.} (good, bad, necessary, expensive, etc.)}\\
    &\mbox{(4) \textbf{Colour.} (green, white, orange, etc.)}
\end{aligned}
\qquad
\begin{aligned}[c] 
    &\mbox{(5) \textbf{Physical Property.} (hard, hot, wet, etc.)}\\
    &\mbox{(6) \textbf{Human Propensity.} (happy, greedy, etc.)}\\
    &\mbox{(7) \textbf{Speed.} (fast, slow, etc.)}\\
    &\mbox{ }\\
\end{aligned}
\end{equation*}

In this paper our focus will be representing nouns other than human beings in conceptual spaces, therefore we will not consider adjectives from item (6). Also, for the purposes of language translation, our focus will be on recreating a human's process of concept construction so the author proposes reframing some of these seven core adjective types from the perspective of our five senses; \textit{sight, smell, sound, sensation and savour}. This paper does not produce examples involving \emph{smell} or \emph{sound}, so these categories have been omitted.

\vspace{-6mm}
\begin{equation*}
\begin{aligned}[c]
    &\mbox{(1) \textbf{Dimension.}}\\
    &\mbox{(2) \textbf{Age.}}\\
    &\mbox{(3) \textbf{Value.}}\\
    &\mbox{(4) \textbf{Speed.}}
\end{aligned}
\qquad\qquad\qquad\qquad
\begin{aligned}[c] 
    &\mbox{(5) \textbf{Physical Property.} Further classified as:}\\
    &\mbox{\hspace{10mm}(a) Colour, Intensity \hspace{52mm}(Sight)}\\
    &\mbox{\hspace{10mm}(b) Savour}\\
    &\mbox{\hspace{10mm}(c) Temperature, Density, Mass, Texture \hspace{15mm}(Sensation)}
\end{aligned}
\end{equation*}

How do we represent this data numerically? Fortunately most aspects of the five categories lend themselves to a linear interpretation. For example, in \textbf{Dimension} we can manually order adjectives in this class from `small' to `large' and represent \textbf{Dimension} as an interval $[0,1]$, assigning adjectives representing `small' sizes values close to 0, and adjectives representing `large' sizes values close to 1. The allocation of these values depends on the preferences of the programmer. This will not be extremely precise - nor, in fact, do we want it to be - by our very nature spaces visualised by humans are fuzzy, and our use of adjectives reflects this. Therefore while there are fuzzy boundaries between sizes within the class, overall there will be a distinction between `small' and `large', though how big that distinction is depends on the programmer inputting these values.

Similarly we allow \textbf{Age} to be represented by $[0,1]$ (where adjectives such as \textit{young, new, baby} are valued closer to $0$, and \textit{old, mature, antiquated} are closer to $1$), and \textbf{Value} and \textbf{Speed} to be represented by $[0,1]$ as well. We will take these spaces as given and assume one can preload a list of common adjectives with assigned $[0,1]$ values, in much the same way it is assumed one can preload a list of colours with assigned $[0,1]^3$ values in the common RGB colour cube. For \textbf{Physical Properties},
\begin{itemize}
    \item[(a)] \textit{Colour} will be represented numerically by the RGB colour cube, and Intensity by the interval $[0,1]$.
    
    \item[(b)] \textit{Savour} by G\"{a}rdenfors' taste tetrahedron (\emph{see} \cite[Fig.~2]{dan}). Embed this into $\mathbb{R}^3$ in the usual way: $\mbox{Salt} = [1, 0, 0]$, $\mbox{Sour} = [-\tfrac{1}{2}, -\tfrac{\sqrt{3}}{2}, 0]$, $\mbox{Bitter} = [-\tfrac{1}{2}, \tfrac{\sqrt{3}}{2}, 0]$, and $\mbox{Sweet} = [0, 0, \sqrt{2}]$. 
    
    \item[(c)] \textit{Sensation} the author suggests representing by a hypercube $[0,1]^4$ with the first dimension temperature (from low to high), the second dimension density (from low - e.g.\ \emph{gaseous, wispy, fine}, to high - e.g.\ \emph{solid, dense, hard}, with items like \emph{soft, mushy, wet, gloopy, sticky, brittle, crumbly} in between), the third dimension mass (from light to heavy) and the fourth dimension texture (from smooth to rough).
\end{itemize}  

This system cannot capture every type of adjective. At present this view is not sophisticated enough to capture `dry', `clear', `sunny' etc. - however, this system does allow us to start analysing text in a meaningful way. Going forward, we shall assign numerical values to adjectives based on our intuition and assume a complex set of adjectives has been hard coded into our algorithm a priori. This may seem a little ad hoc, but it is how we learn adjectives in the early years of our lives: by repeated exposure and memorisation. Of course, our mental picture of objects comes not just from adjectives but also other nouns.

\subsection{Nouns}\label{nouns}
The advantage to allowing nouns to classify other nouns is twofold; first, nouns can identify structure that adjectives might have missed. For example, describing apples and cars as ``red, smooth and fresh smelling'' might be accurate, but paints the wrong conceptual picture. The picture is corrected once we include the sentences ``an apple is a fruit'' and ``a car is a vehicle''. Such classifying words as `fruit' or `vehicle' are known as \emph{hypernyms}; a word $A$ is a hypernym of a word $B$ if the sentence ``$B$ is a (kind of) $A$'' is acceptable to English speakers. The converse, a \emph{hyponym}, is defined as a word $B$ such that the sentence ``$B$ is a (kind of) $A$'' is acceptable. This brings us to the second advantage of allowing nouns into our classification system; like adjectives, they can be ordered (this time in a tree\footnote{It might not be technically correct to refer to the structure as a tree, as each word might have several hypernyms. Nevertheless, the terminology has stuck.}) by the hypernym-hyponym relationship.

There is already a substantial amount of work done on classifying nouns by the hypernym-hyponym relationship, and there exist algorithms which extract this sort of structure from a given corpus \cite{autocorp2, autocorp4, autocorp3, autocorp1}. Hearst \cite{autocorp3} in 1992 revolutionarily algorithmitised hypernym-hyponym relationships according to a certain set of English rules (which, incidentally, can be recreated for Irish). Caraballo \cite{autocorp2} took this work further and produced a working example with the `Wall Street Journal' Penn Treebank corpus \cite{markus}.

As well as this, there already exists the knowledge base WordNet \cite{wordnet} and its Irish counterpart LSG (\textit{L\'{i}onra S\'{e}imeantach na Gaeilge}) \cite{irishwordnet}, both of which have organised thousands of nouns into this hierarchical relationship. Therefore we shall assume a hierarchy such as $\mbox{\textit{food}} \rightarrow \mbox{\textit{fruit}} \rightarrow \mbox{\textit{berry}}$ can already be extracted from text. Using these tools, we have the following options when making use of the hypernym-hyponym tree in concept creation:
\begin{enumerate}
    \item If we are interested solely in concept creation (i.e.\ are only concerned with concepts for \textit{one} language) we can remove the dependency of the tree on the corpus being analysed by using WordNet to create a hypernym-hyponym tree. Relabelling the vertices gives us a convex space associated to each noun in the text via their path from root to leaf. 
    
    \item If we are interested in using concepts for language translation the matter becomes trickier - the trees generated by WordNet and LSG might not have the same structure. However, if we assume we are given two copies of the same corpus, one in English and the other in Irish, then we can assume the \emph{same} (up to synonyms\footnote{Each node in the hypernym-hyponym tree is a `synset' (a class of synonyms).}) hierarchy of nouns is produced in the corresponding languages, using the extraction algorithms created by Hearst and Caraballo (\cite{autocorp3}, \cite{autocorp2}, resp.). 
\end{enumerate}

The key point: given a corpus of text in English producing the hierarchy $\mbox{\textit{food}} \rightarrow \mbox{\textit{fruit}} \rightarrow \mbox{\textit{berry}}$, we will assume the hierarchy $\mbox{\textit{bia}} \rightarrow \mbox{\textit{tortha\'{i}}} \rightarrow \mbox{\textit{caora}}$ produced by the Irish corpus is \emph{directly comparable} to the English hierarchy, meaning we can instead label the hierarchy as $v_0 \rightarrow v_1 \rightarrow v_2$ and refer to berry (and \textit{caora}) by its path in the hierarchy: $\{v_0, v_1, v_2\}$. In \S\ref{bigex} there is an example of this proposal working successfully.

\section{Conceptual Space Creation from a Corpus}\label{five}

In 2004 G\"{a}rdenfors \cite{gardenforce} introduced \emph{conceptual spaces} as a means of representing information in a `human' way; the founding idea being if two objects belong to the same concept, then every object somehow `in between' these objects also belongs to the same concept. We can mathematically describe the property of `in between' via \textit{convex algebras}, an introduction to which is given by Bolt et al.\ \cite[\S 4]{dan}. The work in this section and the next is carried out in the category \textbf{ConvexRel}; the category with convex algebras as objects and convex relations as morphisms. The two convex algebras of interest to us are \emph{Examples 1 \&\ 6} of \cite{dan}:
\begin{enumerate}
    \item The closed real interval $[0,1]$ has a convex algebra structure induced by the vector space $\mathbb{R}$. The formal sums $\sum_i p_i | x_i \rangle$ are sums of elements in $[0,1]$ with addition and multiplication from $\mathbb{R}$. The mixing operation is the identity map.
    
    \item A finite tree can be a convex algebra - in particular, the hypernym-hyponym trees we are interested in are affine semilattices, hence the formal sums
    $$\sum_i p_i | a_i \rangle := \bigvee_i \{a_i \mbox{ : } p_i > 0 \}$$
    are well defined. (So, for example the formal sum $p_1 | x_1 \rangle + p_2 | x_2 \rangle + p_3 | x_3 \rangle$ is the lowest level in the tree containing $x_1, x_2, x_3$; their \emph{join}.)
\end{enumerate}
 \textbf{ConvexRel} is compact closed \cite[Theorem 1]{dan}, hence (by Coecke et al.\ \cite{bob}) combines perfectly with the Lambek grammar category allowing us to create a functor to interpret meanings in the \textbf{ConvexRel} category via the type reductions in the Lambek grammar category.

The first hurdle we must overcome if we wish to use the \textbf{ConvexRel}-DisCoCat machinery is taking words in our foreign language (here Irish) and systematically representing them as convex spaces. The method we propose is reminiscent of how language is learnt in humans. For example, if a friend tells you an \textit{\'{u}ll} is a red, round, smooth, bitter or sweet fruit, you will (eventually, with enough information) come to understand they are describing an \textit{apple}. It is in this vein of thought we present the following definition:

\begin{definition}
A \emph{descriptor} $D$ of a noun $N$ is an adjective or noun which aids in the description of $N$; if $D$ is an adjective it describes physical properties of $N$ (e.g.\ \textit{red, bitter, smooth}) and if $D$ is a noun it classifies $N$ according to nouns in an already-known hierarchical structure (e.g\ \emph{fruit}, belonging to $\mbox{\textit{food}} \rightarrow \mbox{\textit{fruit}} \rightarrow \mbox{\textit{berry}}$).
\end{definition}


The basic idea of conceptual space creation we propose is as follows: given a corpus of text involving heavy use of a noun $N$, parse the text identifying descriptors of $N$. The example corpus, \emph{Corpus \ref{spacecorp}}, is quite simple so this parsing can be achieved by forming a collection of all words occurring in the same sentence as $N$, then sorting this collection into adjectives (which are represented as vectors according to \S \ref{adj}) and nouns (which are represented by a tree according to \S\ref{nouns}). Taking the convex hull of the points in each adjective type, then the tensor product of the convex hulls, we represent the adjective descriptors of $N$ as a convex set. Noun descriptors are represented as a convex set a l\`{a} \S\ref{nouns}. Combining these convex subsets under a tensor product gives us a conceptual space, as required.

\subsection{Example: Planets, the Sun and Fruit.}\label{bigex}
Consider \emph{Corpus \ref{spacecorp}} from \emph{Appendix \ref{spaceappendix}}. Let us examine five main nouns from this corpus; 
$$N^1 = \mbox{Venus, } N^2 = \mbox{Jupiter, } N^3 = \mbox{Mars, } N^4 = \mbox{apple, } N^5 = \mbox{The Sun}.$$
Organising this into a table we obtain:

\begin{center}
\begin{longtable}{ |c|c|p{9cm}| } 
 \hline
 Venus & \textbf{Adjectives} & solid, rocky, same size as Earth, hot, high pressure, bright.\\
 \hline
 & \textbf{Nouns} & planet, Earth's sister, ball.\\
 \hline
  Jupiter & \textbf{Adjectives} & very large, gassy, orange, brown, red, far away, windy, freezing, very bright.\\
 \hline
 & \textbf{Nouns} & planet, ball, outer space.\\
 \hline
  Mars & \textbf{Adjectives} & very red, brown, orange, cold, smaller than Earth, rocky.\\
 \hline
 & \textbf{Nouns} & planet, outer space.\\
 \hline
  Apple & \textbf{Adjectives} & round, soft, red, green, bitter, sweet.\\
 \hline
 & \textbf{Nouns} & fruit, ball.\\
 \hline
  The Sun & \textbf{Adjectives} & brightest, huge, very hot, round, very dense. \\
 \hline
 & \textbf{Nouns} & star, ball.\\
 \hline
\end{longtable}
\end{center}
\vspace{-6mm}
We first deal with the adjectives.  These can be organised according to \emph{Section \ref{adj}}:

\noindent(1) \textbf{Venus.}
\vspace{-3mm}
{\small\begin{align*}
&N_{\mbox{\tiny  dimension}} = Conv(\mbox{same size as Earth}) = \{0.5\},\\
&N_{\mbox{\tiny  intensity}} = Conv(\mbox{bright}) = \{0.7\},\\
&N_{\mbox{\tiny  temperature}} = Conv(\mbox{hot}) = \{0.75\},\\
&N_{\mbox{\tiny  density}} = Conv(\mbox{solid}) = \{0.9\},\\
&N_{\mbox{\tiny  texture}} = Conv(\mbox{rocky}) = \{0.9\}.
\end{align*}}

\vspace{-5mm}
(These values were assigned according to the author's own preference, however they can be assigned different values according to each readers' wishes.) $D^1_{\mbox{\tiny  adj}}$ is the tensor product of these spaces, where if an adjective class is not mentioned, its corresponding noun space (or \emph{property}) is set to $[0,1]$ (e.g.\ $N_{\mbox{\tiny age}} = [0,1]$). Note that we were required to drop some adjectives, such as \emph{high pressure}, as our adjective classification from \emph{Section \ref{adj}} is not specific enough to capture all details. Also note that it is also unusual that these spaces are singleton sets; in a larger, more complicated corpus these properties would be intervals.

\vspace{1mm}{\small
\hspace{-4mm}\begin{minipage}{0.45\textwidth}
\vspace{-5mm}\begin{align*}
\mbox{(2) }&\mbox{\textbf{Jupiter.}}\\
&N_{\mbox{\tiny  dimension}} = Conv(\mbox{very large}) = \{0.7\},\\
&N_{\mbox{\tiny  colour}} = Conv(\mbox{orange} \cup \mbox{brown} \cup \mbox{red})\footnotemark,\\
&N_{\mbox{\tiny  intensity}} = Conv(\mbox{very bright}) = \{0.8\},\\
&N_{\mbox{\tiny  temperature}} = Conv(\mbox{freezing}) = \{0\},\\
&N_{\mbox{\tiny  density}} = Conv(\mbox{gassy}) = \{0.1\}.
\end{align*}
\end{minipage}\hspace{9mm}\begin{minipage}{0.45\textwidth}
\begin{align*}
\mbox{(3) }&\mbox{\textbf{Mars.}}\\
&N_{\mbox{\tiny  dimension}} = Conv(\mbox{smaller than Earth}) = \{0.25\},\\
&N_{\mbox{\tiny  colour}} = Conv(\mbox{red} \cup \mbox{brown} \cup \mbox{orange}),\\
&N_{\mbox{\tiny  temperature}} = Conv(\mbox{cold}) = \{0.4\},\\
&N_{\mbox{\tiny  texture}} = Conv(\mbox{rocky}) = \{0.9\}.
&\mbox{ }\\
&\mbox{ }\\
\end{align*}
\end{minipage}}
\footnotetext{The RGB values we use for \emph{orange} and \emph{brown }are (255,165,0) and (153,76,0) respectively.}

{\small
\hspace{-4mm}\begin{minipage}{0.45\textwidth}
\vspace{-5mm}\begin{align*}
\mbox{(4) }&\mbox{\textbf{Apple.}}\\
&N_{\mbox{\tiny  colour}} = Conv(\mbox{red} \cup \mbox{green}),\\
&N_{\mbox{\tiny  taste}} = Conv(\mbox{bitter} \cup \mbox{sweet}),\\
&N_{\mbox{\tiny  texture}} = Conv(\mbox{soft}) = \{0.4\}.
&\mbox{ }\\
\end{align*}
\end{minipage}\hspace{0mm}\begin{minipage}{0.45\textwidth}
\vspace{-5mm}\begin{align*}
\mbox{(5) }&\mbox{\textbf{The Sun.}}\\
&N_{\mbox{\tiny  dimension}} = Conv(\mbox{huge}) = \{1\},\\
&N_{\mbox{\tiny  intensity}} = Conv(\mbox{brightest}) = \{1\},\\
&N_{\mbox{\tiny  temperature}} = Conv(\mbox{very hot}) = \{1\},\\
&N_{\mbox{\tiny  density}} = Conv(\mbox{very dense}) = \{1\}.
\end{align*}
\end{minipage}}
\vspace{2.5mm}

For $K = 2,3,4,5$, $D^K_{\mbox{\tiny  adj}}$ is the tensor product of the $N_{\mbox{\scriptsize(--)}}$ spaces of item (K). The additional information of the noun interdependence is added by the following tree, generated by WordNet \cite{wordnet}: (see the tree on the following page).

The underlined leaves are those nouns appearing as descriptors in \emph{Corpus \ref{spacecorp}}. Referring to each node by its $e_i$ label, we can define the sets $D^i_{\mbox{\tiny noun}}$.

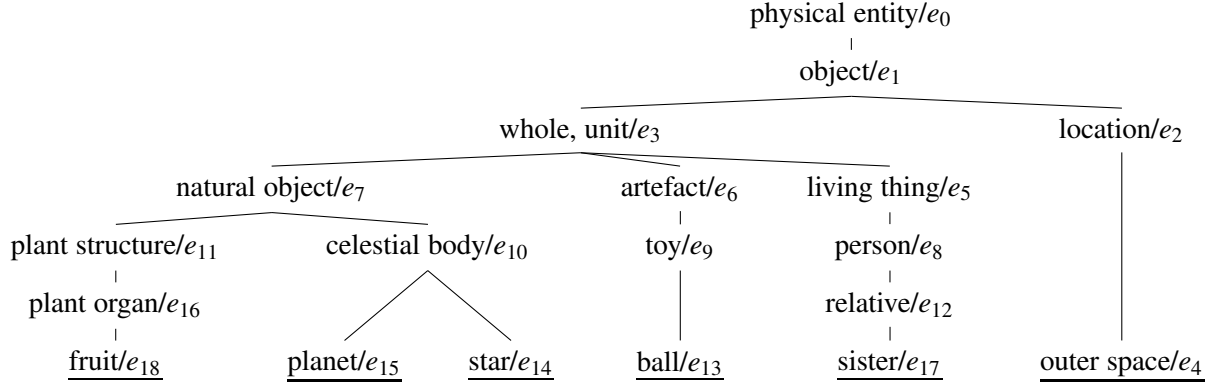
\vspace{-6mm}
\begin{center}
\begin{tikzpicture}
\begin{scope}[frontier/.style={distance from root=132pt}]
\tikzset{level distance=22pt,sibling distance=18pt}
\Tree [.{physical entity/$e_0$} [.object/$e_1$ [.{whole, unit/$e_3$} [.{natural object/$e_7$} [.{plant structure/$e_{11}$} [.{plant organ/$e_{16}$} \node(fruit){\underline{fruit/$e_{18}$}}; ] ] [.{celestial body/$e_{10}$} \node(planet){\underline{planet/$e_{15}$}}; \node(star){\underline{star/$e_{14}$}}; ] ] [.artefact/$e_6$ [.toy/$e_9$ \node(ball){\underline{ball/$e_{13}$}}; ] ] [.{living thing/$e_5$} [.person/$e_8$ [.relative/$e_{12}$  \node(sister){\underline{sister/$e_{17}$}}; ] ] ] ] [.location/$e_2$ \node(space){\underline{outer space/$e_4$}}; ] ] ]
\end{scope}
\end{tikzpicture}    
\end{center}

\begin{minipage}{0.5\textwidth}
\begin{align*}
&D^1_{\mbox{\tiny  noun}} \defeq \{e_0, e_1, e_3, e_5, e_6, e_7, e_8, e_9, e_{10}, e_{12}, e_{13}, e_{15}, e_{17}\},\\
&D^2_{\mbox{\tiny  noun}} \defeq \{e_0, e_1, e_2, e_3, e_4, e_6, e_7, e_9, e_{10}, e_{13}, e_{15}\},\\
&D^3_{\mbox{\tiny  noun}} \defeq \{e_0, e_1, e_2, e_3, e_4, e_7, e_{10}, e_{15}\},\\
&D^4_{\mbox{\tiny  noun}} \defeq \{e_0, e_1, e_3, e_6, e_7, e_9, e_{11}, e_{13}, e_{16}, e_{18}\},\mbox{\hspace{10mm}}\\
&D^5_{\mbox{\tiny  noun}} \defeq \{e_0, e_1, e_3, e_6, e_7, e_9, e_{10}, e_{13}, e_{14}\},
\end{align*}
\end{minipage}
\begin{minipage}{0.5\textwidth}
\begin{align*}
&\mbox{Venus} := D^1_{\mbox{\tiny  adj}} \otimes D^1_{\mbox{\tiny  noun}},\\
&\mbox{Jupiter} := D^2_{\mbox{\tiny  adj}} \otimes D^2_{\mbox{\tiny  noun}},\\
&\mbox{Mars} := D^3_{\mbox{\tiny  adj}} \otimes D^3_{\mbox{\tiny  noun}},\\
\mbox{\hspace{13mm}}&\mbox{Apple} := D^4_{\mbox{\tiny  adj}} \otimes D^4_{\mbox{\tiny  noun}},\\
&\mbox{The Sun} := D^5_{\mbox{\tiny  adj}} \otimes D^5_{\mbox{\tiny  noun}}.
\end{align*}
\end{minipage}
\vspace{2mm}

\dots and finally we obtain the conceptual spaces for the 5 nouns, on the right.

In Irish, the same corpus is \emph{Corpus \ref{spacecorpirish}}, located in \emph{Appendix \ref{spaceappendix}}. The five main nouns of this corpus are (in no particular order) $M^1 = \mbox{\it V\'{e}ineas, } M^2 = \mbox{\it I\'{u}patar, } M^3 = \mbox{\it Mars, } M^4 = \mbox{\it \'{U}ll, } M^5 = \mbox{\it Grian}$.

We can once again organise the information of \emph{Corpus \ref{spacecorpirish}} into a table (see \emph{Appendix \ref{rename}}) and determine noun spaces $N_{\mbox{\scriptsize(--)}}$ from adjectives as was explained in \S\ref{adj}. For example, in the case of \emph{V\'{e}ineas:}

\vspace{-6mm}
\begin{align*}
&N_{\mbox{\tiny  dimension}} = Conv(\mbox{\it {m\'{e}id c\'{e}anna leis an Domhan}}) = \{0.5\},\\
&N_{\mbox{\tiny  intensity}} = Conv(\mbox{\it {geal}}) = \{0.6\},\\
&N_{\mbox{\tiny  temperature}} = Conv(\mbox{\it {an-te}}) = \{0.85\},\\
&N_{\mbox{\tiny  density}} = Conv(\mbox{\it {tathagach}}) = \{0.9\},\\
&N_{\mbox{\tiny  texture}} = Conv(\mbox{\it {carraigeach}}) = \{0.9\}.
\end{align*}

Note that the values here are different than the corresponding values in English for \textit{geal} (bright), \textit{an-te} (hot), etc. The reasoning here is as follows: in Irish there is no word for ``hot'' - to describe high temperatures there is just ``warm'' and ``very warm''. So ``\textit{an-te}'' (``very warm'') suffices for ``hot'', therefore since ``\textit{an-te}'' is the hottest the weather can be described, it is assigned a value of 0.85 in Irish (because in English, ``very hot'' would need to correspond to a higher value than ``hot'', which is 0.75). 

$\overline{D}^1_{\mbox{\tiny adj}}$ we define to be the tensor product of the $N_{\mbox{\scriptsize(--)}}$ spaces of \emph{V\'{e}ineas}. Note that we were required to drop some adjectives, such as \textit{{br\'{u}\dots ard}} (\textit{high pressure}), as our adjective classification from \S\ref{adj} is not specific enough to capture all details.

The additional linguistic information from the descriptor nouns is obtained by referencing a hypernym-hyponym tree, which for example WordNet (in Irish) organises as:

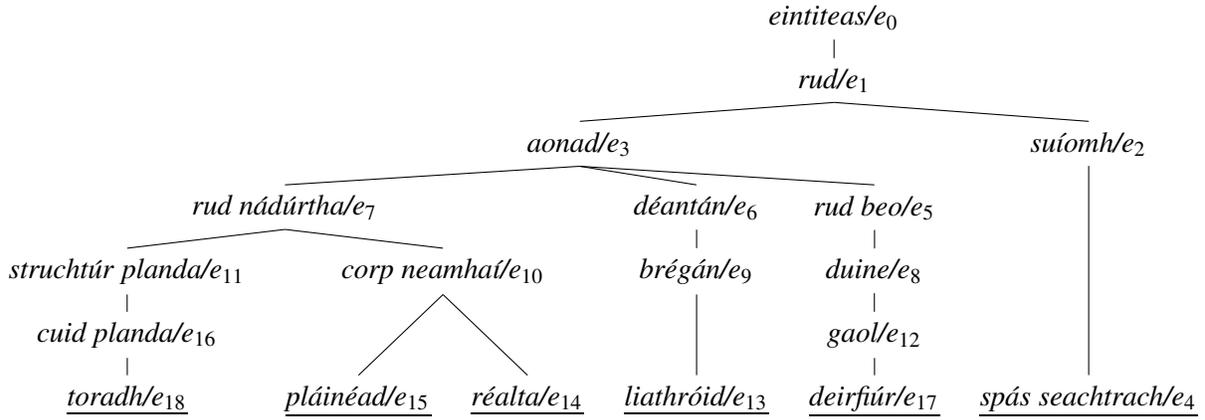
\vspace{-5mm}
\begin{center}
\begin{tikzpicture}
\begin{scope}[frontier/.style={distance from root=144pt}]
\tikzset{level distance=24pt,sibling distance=8pt}
\Tree [.{\it eintiteas/$e_0$} [.{\it rud/$e_1$} [.{\it {aonad/$e_3$}} [.{\it {rud n\'{a}d\'{u}rtha/$e_7$}} [.{\it strucht\'{u}r planda/$e_{11}$} [.{\it cuid planda/$e_{16}$} \node(fruit){\underline{\it toradh/$e_{18}$}}; ] ] [.{\it corp neamha\'{i}/$e_{10}$} \node(planet){\it \underline{pl\'{a}in\'{e}ad/$e_{15}$}}; \node(star){\it \underline{r\'{e}alta/$e_{14}$}}; ] ] [.{\it d\'{e}ant\'{a}n/$e_6$} [.{\it br\'{e}g\'{a}n/$e_9$} \node(ball){\underline{\it {liathr\'{o}id/$e_{13}$}}}; ] ] [.{\it {rud beo/$e_5$}} [.{\it duine/$e_8$} [.{\it gaol/$e_{12}$}  \node(sister){\underline{\it {deirfi\'{u}r/$e_{17}$}}}; ] ] ] ] [.{\it su\'{i}omh/$e_2$} \node(space){\underline{\it {sp\'{a}s seachtrach/$e_4$}}}; ] ] ]
\end{scope}
\end{tikzpicture}    
\end{center}

If we label the tree counterpart to the English tree, we can finally define the conceptual spaces for the 5 main nouns. 

\begin{minipage}{0.5\textwidth}
\begin{align*}
&\overline{D}^1_{\mbox{\tiny  noun}} \defeq \{e_0, e_1, e_3, e_5, e_6, e_7, e_8, e_9, e_{10}, e_{12}, e_{13}, e_{15}, e_{17}\},\\
&\overline{D}^2_{\mbox{\tiny  noun}} \defeq \{e_0, e_1, e_2, e_3, e_4, e_6, e_7, e_9, e_{10}, e_{13}, e_{15}\},\\
&\overline{D}^3_{\mbox{\tiny  noun}} \defeq \{e_0, e_1, e_2, e_3, e_4,  e_7, e_{10}, e_{15}\},\\
&\overline{D}^4_{\mbox{\tiny  noun}} \defeq \{e_0, e_1, e_3, e_6, e_7, e_9, e_{11}, e_{13}, e_{16}, e_{18}\},\\
&\overline{D}^5_{\mbox{\tiny  noun}} \defeq \{e_0, e_1, e_3, e_6, e_7, e_9, e_{10}, e_{13}, e_{14}\}.
\end{align*}
\end{minipage}
\begin{minipage}{0.5\textwidth}
\begin{align*}
&\mbox{\it V\'{e}ineas} := \overline{D}^1_{\mbox{\tiny  adj}} \otimes \overline{D}^1_{\mbox{\tiny  noun}},\\
&\mbox{\it I\'{u}patar} := \overline{D}^2_{\mbox{\tiny  adj}} \otimes \overline{D}^2_{\mbox{\tiny  noun}},\\
&\mbox{\it Mars} := \overline{D}^3_{\mbox{\tiny  adj}} \otimes \overline{D}^3_{\mbox{\tiny  noun}},\\
&\mbox{\it \'{U}ll} := \overline{D}^4_{\mbox{\tiny  adj}} \otimes \overline{D}^4_{\mbox{\tiny  noun}},\\
&\mbox{\it Grian} := \overline{D}^5_{\mbox{\tiny  adj}} \otimes \overline{D}^5_{\mbox{\tiny  noun}}.
\end{align*}
\end{minipage}

\bigskip
\section{Metrics for concepts}\label{7}
Our final goal is to compare the concepts created in \S\ref{bigex} in Irish and English. To do this we require some measure of distance between concepts; we require a metric on \textbf{ConvexRel}. We will define the metric $d$ then leave the technical details concerning the combination of convex structures and metrics to \cite{hypergraphs} (cf.\ \emph{Example 28}, ibid.).

\begin{definition}
\textbf{(Hausdorff Metric). }Let $X, Y$ be two nonempty subsets of a metric space $(M, f)$. Define their \emph{Hausdorff distance} to be
$$d(X, Y) \defeq \max\{ \sup_{x \in X} \inf_{y \in Y} f(x, y)\mbox{, } \sup_{y \in Y} \inf_{x \in X} f(x, y)\}.$$
\end{definition}

In the case of \textbf{ConvexRel}, all the concepts we define are subsets of $\mathbb{R}^{20} \otimes E$ where $E = \{e_0, \dots, e_n\}$ represents a tree, according to \S \ref{nouns}. On $\mathbb{R}^{20}$ there is the taxicab metric and for $E$ there is the metric $f$ on $\mathcal{P}(E)$, the power set of $E$, defined by
$$\mbox{for }A, B \subseteq E\mbox{,} \quad f(A, B) \defeq \max\{|A \setminus B|\mbox{, } |B \setminus A|\}.$$

\begin{example}
Consider the distance between ``Apple'' and ``Jupiter'', whose conceptual spaces were calculated in \S \ref{bigex}. 
\newpage
{\small\begin{align*}
&d(\mbox{``Apple''}, \mbox{``Jupiter''}) \quad= \quad \max\{ \sup_{x \in \mbox{\tiny Apple}} \inf_{y \in \mbox{\tiny Jupiter}} f(x, y)\mbox{, } \sup_{y \in \mbox{\tiny Jupiter}} \inf_{x \in \mbox{\tiny Apple}} f(x, y)\}\\
&= \max\{\sup_{x \in \mbox{\tiny Apple}} \inf_{y \in \mbox{\tiny Jupiter}} (f(N^{\mbox{\tiny Apple}}_{\mbox{\tiny dimension}},N^{\mbox{\tiny Jupiter}}_{\mbox{\tiny dimension}}) + f(N^{\mbox{\tiny Apple}}_{\mbox{\tiny colour}},N^{\mbox{\tiny Jupiter}}_{\mbox{\tiny colour}}) +\dots + f(D^{\mbox{\tiny Apple}}_{\mbox{\tiny noun}},\overline{D}^{\mbox{\tiny Jupiter}}_{\mbox{\tiny noun}}))\mbox{, } \inf_{x \in \mbox{\tiny Apple}} \sup_{y \in \mbox{\tiny Jupiter}} \dots\}\\
&= \max\{\sup_{x \in \mbox{\tiny Apple}} \inf_{y \in \mbox{\tiny Jupiter}} (f([0,1], \{0.7\}) + f(Conv(\mbox{red}\cup\mbox{green}), Conv(\mbox{orange}\cup\mbox{brown}\cup\mbox{red}) + \dots \\
&\hspace{13mm}+ f([0,1], \{0.1\}) + f(\{e_0, e_1, e_3, e_6, e_7, e_9, e_{11}, e_{13}, e_{16}, e_{18}\},\{e_0, e_1, e_2, e_3, e_4, e_6, e_7, e_9, e_{10}, e_{13}, e_{15}\}), \dots\}\\
&= 8.7.
\end{align*}}
Similarly, \vspace{-3mm}
\begin{align*}
 d(\mbox{``Mars''}, \mbox{``Jupiter''}) &= 5.55,\\
 d(\mbox{``Jupiter''}, \mbox{``Sun''}) &= 7.7,\\
 d(\mbox{``Apple''}, \mbox{``Sun''}) &= 7.97.
\end{align*}

This seems to capture the rough picture we desire: relatively speaking\footnote{Relative to the other distances calculated in this example. 
}, the planets Mars and Jupiter are close, while nouns like ``Apple'' and ``Jupiter'' or ``Apple'' and ``Sun'' are distant. ``Sun'' is also technically closer to ``Jupiter'' than to ``Apple'', though not by much. One might expect ``Jupiter'' to be closer to ``Sun'', however this is not the picture \emph{Corpus \ref{spacecorp}} paints; in it, the Sun is not a planet and is described as ``very hot'' or ``very dense''. Perhaps if the corpus noted the Sun is `planet-like' or `round like a planet', and described the colour of the sun as `yellow and orange', the distance between these two concepts might be smaller.
\end{example}

Finally, let us return to translation between Irish and English. Using the same metric, the distance between ``Apple'' and its Irish translation, ``\textit{\'{U}ll}'', is given by $d(\mbox{``Apple''}, \mbox{``\textit{\'{U}ll}''}) = 0$, which is to say as concepts, ``Apple'' and ``\textit{\'{U}ll}'' are equal. On the other hand, the distance between ``Apple'' and ``\textit{Grian}'' (English: ``Sun'') is $d(\mbox{``Apple''}, \mbox{``\textit{Grian}''}) = 7.97$.

If we were to attempt to translate ``\textit{I\'{u}patar}'' using the metric on \textbf{ConvexRel}, we see

\vspace{-1mm}
\begin{equation*}
\begin{aligned}[c]
d(\mbox{``Venus''}, \mbox{``\textit{I\'{u}patar}''}) &= 8.75,\\
d(\mbox{``Jupiter''}, \mbox{``\textit{I\'{u}patar}''}) &= 0.3,\footnotemark\\
d(\mbox{``Mars''}, \mbox{``\textit{I\'{u}patar}''}) &= 5.45.
\end{aligned}
\qquad
\begin{aligned}[c]
d(\mbox{``Apple''}, \mbox{``\textit{I\'{u}patar}''}) &= 8.6,\\
d(\mbox{``Sun''}, \mbox{``\textit{I\'{u}patar}''}) &= 7.6,\\
\end{aligned}
\end{equation*}
\footnotetext{That this is non-zero stems from the fact that adjectives can have different meanings with different intensities in different languages.}

\vspace{-2mm}
Hence choosing the concept closest to ``\textit{I\'{u}patar}'', which is ``Jupiter'', we deduce we have indeed successfully translated this word.

\begin{remark}
\emph{It is the opinion of the author that the exercise of manually inputting values for the seven core adjective types is an important, maybe even necessary, one. This method is how we first master colours and smells and sizes; by hearing about them and memorising terms, ordered relative to each other. In the words of G\"{a}rdenfors \cite{gardenforce2}, ``we are not born with our concepts; they must be learned''.
The author believes it is also necessary to preform this exercise separately for Irish, as adjectives in this language can have different emphases and occasionally different meanings.}
\end{remark}

\section{Conclusion}

This paper has outlined two methods of translating from Irish to English using the distributional compositional categorical model of meaning; via vector spaces and the category \textbf{FVect} and via concepts and the category \textbf{ConvexRel}. The former allowed us to compare the meanings of sentences between languages by calculating similarity scores, and the latter allowed us to focus more on the meaning behind nouns and calculate distances between concepts across languages. 

The work of this paper can be extended in many ways. In \emph{Section \ref{irishlambek}}, the Lambek pregroup grammar structure for Irish can be further embellished and more grammatical features of Irish captured as Lambek does for English \cite{lambek}. The ideas behind \emph{Section \ref{claso}} can also be expanded to address adjectives not captured by the system presented in \emph{Section \ref{adj}} and \emph{Section \ref{five}}, titled `Conceptual Space Creation from a Corpus', could also be computationally tested with larger corpora. In particular, a more precise explanation and demonstration of how the descriptors of a noun $N$ are identified and sorted for any large corpus could be further addressed.

Ongoing work includes determining a treatment for quantification and negation, often important for language translation. Also, perhaps most importantly, the setting proposed in this paper must be implemented, evaluated, and experimented with to produce a useful tool for understanding Irish.

\bigskip
\section*{Acknowledgements}
The author would like to extend his thanks to Professor Coecke and Dr.\ Marsden for their help in this topic throughout the year. The author's thanks to Kevin Scannell for his communication and for providing the source files to the LSG. The author would also like to thank Abby Breathnach and Soinbhe Nic Dhonncha for their help in translating key examples in this paper to Irish. Finally, the author thanks Martha Lewis and the CAPNS referees for their helpful comments and suggestions in their review of this paper.

\bibliographystyle{eptcs}
\bibliography{papers}

\begin{appendices}
\section{Corpus for Vector Space Model of Meaning (English)}\label{sweng}
The following is a summary of \textit{Star Wars: Episode III - Revenge of the Sith}, obtained from Wikipedia \cite{wiki} and edited by the author. Note that we are making some assumptions in using this corpus. The author is assuming the model of meaning can understand third-person sentences as if they were first-person sentences; i.e.\ ``she is pregnant'' is understood to be ``Padm\'{e} is pregnant''. We are also assuming the model can understand sentences with conjunction; e.g.\ ``Anakin and Obi-Wan are known for their bravery'' is ``Anakin is known for his bravery'' and ``Obi-Wan is known for his bravery''. We assume the model can understand the use of the present participle, i.e.\ ``After infiltrating General Grevious' flagship'' is understood to be ``After Anakin and Obi-Wan infiltrate General Grevious' flagship''. Finally we also assume the corpus has been lemmatised for \emph{Sections \ref{vect} \&\ \ref{irishvect}}.

It is true that some of these assumptions might be difficult to work into the vector space model of meaning, however the author feels the use of this corpus gives good examples in \emph{Sections \ref{vect} \&\ \ref{irishvect}} while still being interesting for humans to parse. \emph{Corpora \ref{corpa1} \&\ \ref{corpb1}} can be rewritten such that the above assumptions are no longer necessary, however the story becomes tedious to read.

\begin{corpus}\label{corpa1}
\underline{Palpatine is a mastermind who turns Anakin to the dark side of the Force.}\\

The galaxy is in a state of civil war. Jedi Knights Obi-Wan Kenobi and Anakin Skywalker lead a mission to rescue the kidnapped Supreme Chancellor Palpatine from the evil General Grievous, who is a Seperatist commander. Anakin and Obi-Wan are known for their bravery and skill. After infiltrating General Grievous's flagship, the Jedi duel Dooku, whom Anakin eventually executes at Palpatine's urging. General Grievous escapes the battle-torn cruiser, in which the Jedi crash-land on Coruscant. There Anakin reunites with his beautiful wife, Padm\'{e} Amidala, who reveals that she is pregnant. While initially excited, the prophetic visions that Anakin has cause him to worry. Anakin believes Padm\'{e} will die in childbirth.

Palpatine appoints Anakin to the Jedi Council as his representative. The Jedi do not trust Palpatine as they believe he is too powerful. The Council orders Anakin to spy on Palpatine, his friend. Anakin begins to turn away from the Jedi because of this. Meanwhile the Jedi are searching for a Sith Lord. A Sith Lord is evil person who uses the dark side of the Force. The Jedi try prevent anyone from turning to the dark side of the Force and to evil. Palpatine tempts Anakin with secret knowledge of the dark side of the Force, including the power to save his loved ones from dying. Meanwhile, the powerful Obi-Wan travels to confront General Grievous. The Jedi and General Grievous duel and Obi-Wan fights bravely. Obi-Wan wins his duel against General Grievous. The Jedi Yoda travels to Kashyyyk to defend the planet from invasion. The mastermind Palpatine eventually reveals that he is a powerful Sith Lord to Anakin. Palpatine claims only he has the knowledge to save Padm\'{e} from death. Anakin turns away from Palpatine and reports Palpatine's evil to the Jedi Mace Windu. Mace Windu then bravely confronts Palpatine, severely disfiguring him in the process. Fearing that he will lose Padm\'{e}, Anakin intervenes. Anakin is a powerful Jedi and he severs Mace Windu's hand. This distraction allows Palpatine to throw Mace Windu out of a window to his death. Anakin turns himself to the dark side of the Force and to Palpatine, who dubs him Darth Vader. Palpatine issues Order 66 for the clone troopers to kill the remaining Jedi, then dispatches Anakin with a band of clones to kill everyone in the Jedi Temple. Anakin ventures to Mustafar and massacres the remaining Separatist leaders hiding on the volcanic planet, while Palpatine addresses the Galactic Senate. He transforms the Republic into the Galactic Empire and declares himself Emperor Palpatine.

Obi-Wan and Yoda return to Coruscant and learn of Anakin's betrayal against them. Obi-Wan leaves to talk to Padm\'{e}. Obi-Wan tries to convince her that Anakin has turned to the dark side of the Force; that Anakin has turned to evil. A brave Padm\'{e} travels to Mustafar and implores Anakin to abandon the dark side of the Force. Anakin refuses to stop using the dark side of the Force and sees Obi-Wan hiding on Padm\'{e}’s ship. Anakin angrily chokes Padm\'{e} into unconsciousness. Obi-Wan duels and defeats Anakin. Obi-Wan severs both of his legs and leaves him at the bank of a lava river where he is horribly burned. Yoda duels Emperor Palpatine on Coruscant until their battle reaches a stalemate. Yoda is a powerful Jedi, but he cannot defeat the evil Emperor Palpatine. Yoda then flees with Bail Organa while Palpatine travels to Mustafar. Evil Emperor Palpatine uses the dark side of the Force to sense Anakin is in danger.

Obi-Wan turns to Yoda to regroup. Padm\'{e} gives birth to a twin son and daughter whom she names Luke and Leia. Padm\'{e} dies of sadness shortly after. Palpatine finds a horribly burnt Anakin still alive on Mustafar. After returning to Coruscant, Anakin’s mutilated body is treated and covered in a black armored suit. Palpatine lies to Anakin that he killed Padm\'{e} in his rage. Palpatine is an evil mastermind and leaves Anakin feeling devastated. Palpatine has won; the dark side of the Force now flows through Anakin. Meanwhile, Obi-Wan and Yoda work to conceal the twins from the dark side of the Force, because the twins are the galaxy's only hope for freedom. Yoda exiles himself to the planet Dagobah, while Anakin and the Emperor Palpatine oversee the construction of the Death Star. Bail Organa adopts Leia and takes her to Alderaan. Obi-Wan travels with Luke to Tatooine. There Obi-Wan intends to bravely watch over Luke and his step-family until the time is right to challenge the Empire.
\end{corpus}

\section{Corpus for Vector Space Model of Meaning (Irish)}

For the sake of completeness we give the full Irish corpus whose translated meaning replicates \emph{Corpus \ref{corpa1}}.

\begin{corpus}\label{corpb1}
{
\underline{Is ceannm\'{a}istir a casann Anakin go taobh dorcha na F\'{o}rsa \'{e} Palpatine.}\\

T\'{a} an r\'{e}altra i st\'{a}t cogaidh shibhialta. Rinne Ridir\'{i} Jedi Obi-Wan Kenobi agus Anakin Skywalker misean chun an Seansail\'{e}ir Uachtarach Palpatine a sh\'{a}bh\'{a}il \'{o}n Ginear\'{a}l Grievous olc, ceannasa\'{i} Seperatist \'{e}. Aithn\'{i}tear Anakin agus Obi-Wan d\'{a} a gcr\'{o}gacht agus d\'{a} scileanna. Tar \'{e}is longcheannais Ginear\'{a}l Grievous a ions\'{i}othl\'{a}it, troid na Jedi le Dooku, a mhora\'{i}onn Anakin ar deireadh thiar ar mholadh Palpatine. \'{E}ala\'{i}onn an Ginear\'{a}l Grievous \'{o}n t-\'{e}adromaire caithe, ina dturling\'{i}onn na Jedi chun talamh Coruscant. Ansin, tagann Anakin le ch\'{e}ile lena bhean \'{a}lainn, Padm\'{e} Amidala, a l\'{e}ir\'{i}onn go bhfuil s\'{i} ag iompar clainne. C\'{e} go bhfuil Anakin ar b\'{i}s ar dt\'{u}s, tugann a fh\'{i}seanna f\'{a}idhi\'{u}la c\'{u}is imn\'{i} d\'{o}. Creideann Anakin go gheobhaidh Padm\'{e} b\'{a}s i mbreithe clainne. 

Ceapann Palpatine Anakin chuig Chomhairle na Jedi mar ionada\'{i}. N\'{i}l muin\'{i}n ag na Jedi i Palpatine mar a chreideann siad go bhfuil s\'{e} r\'{o}-chumhachtach. D'orda\'{i}onn an Chomhairle Anakin a dh\'{e}anann spiaireacht ar Palpatine, a chara. Casann Anakin as an Jedi as seo. Idir an d\'{a} linn t\'{a} na Jedi ag cuardach do Tiarna Sith. Is duine olc \'{e} Tiarna Sith a \'{u}s\'{a}ideann an taobh dorcha den Fh\'{o}rsa. D\'{e}anann na Jedi iarracht a chur ar dhuine ar bith a bheith ag casadh go taobh dorcha na F\'{o}rsa agus go holc. Taca\'{i}onn Palpatine Anakin le heolas r\'{u}nda ar thaobh dorcha na F\'{o}rsa, lena n-\'{a}ir\'{i}tear an chumhacht chun a mhuintir a sh\'{a}bh\'{a}il \'{o} bh\'{a}s. Idir an d\'{a} linn, t\'{e}ann Obi-Wan cumhachtach chun dul i ngleic leis an Ginear\'{a}l Grievous. Troideann an Jedi agus Ginear\'{a}l Grievous agus t\'{a} Obi-Wan ag troid go crua. Buaileann Obi-Wan a chath i gcoinne Ginear\'{a}l Grievous. T\'{e}ann Jedi Yoda go Kashyyyk chun an phl\'{a}in\'{e}id a chosaint \'{o} ionradh. L\'{e}ir\'{i}onn an ceannm\'{a}istir Palpatine sa deireadh gurb \'{e} Tiarna cumhachtach Sith \'{e} go Anakin. \'{E}il\'{i}onn Palpatine ach go bhfuil eolas air amh\'{a}in Padm\'{e} a sh\'{a}bh\'{a}il \'{o}n mb\'{a}s. Casann Anakin i gcoinne Palpatine agus tuairisc\'{i}onn s\'{e} olc Palpatine chuig an Jedi Mace Windu. Tabhair Mace Windu aghaidh cr\'{o}ga ar Palpatine, agus \'{e} a dh\'{i}shealbh\'{u} go m\'{o}r sa phr\'{o}iseas. Ag eagla go gcaillfidh s\'{e} Padm\'{e}, idirghabhann Anakin. Is Jedi cumhachtach \'{e} Anakin agus seala\'{i}onn s\'{e} l\'{a}mh Mace Windu. Tugann an t-imr\'{e}iteach seo do Palpatine Mace Windu a chaitheamh as fuinneog go dt\'{i} a bh\'{a}s. Casann Anakin f\'{e}in go taobh dhorcha na F\'{o}rsa agus chuig Palpatine, a ainm Darth Vader d\'{o}. Eis\'{i}onn Palpatine Ord\'{u} 66 do na tr\'{u}pa\'{i} cl\'{o}n chun na Jedi at\'{a} f\'{a}gtha a mhar\'{u}, agus ansin cuireann s\'{e} Anakin le banna cluain\'{e} chuig an Teampaill Jedi a chuir b\'{a}s ar gach duine. Taistil\'{i}onn Anakin go Mustafar agus mais\'{i}onn na ceannair\'{i} Separatist at\'{a} f\'{a}gtha i bhfolach ar an phl\'{a}in\'{e}id b\'{o}lcanach, agus tugann Palpatine aitheasc don Seanad R\'{e}altrach. Athra\'{i}onn s\'{e} an Poblacht isteach sa Impireacht R\'{e}altrach agus dearbha\'{i}onn s\'{e} f\'{e}in an t-Impire Palpatine.

F\'{a}gann Obi-Wan agus Yoda go Coruscant agus foghlaim\'{i}onn siad brad\'{u} Anakin i gcoinne iad. F\'{a}gann Obi-Wan chun labhairt le Padm\'{e}. D\'{e}anann Obi-Wan iarracht a chur ina lu\'{i} di go bhfuil Anakin tar \'{e}is casadh go taobh dorcha na F\'{o}rsa; go bhfuil Anakin tar \'{e}is casadh go holc. Taisteala\'{i}onn Padm\'{e} cr\'{o}ga go Mustafar agus cuireann s\'{i} ar Anakin an taobh dorcha den Fh\'{o}rsa a thr\'{e}igean. Di\'{u}lta\'{i}onn Anakin gan stop a bhaint as an taobh dorcha den Fh\'{o}rsa agus feiceann s\'{e} Obi-Wan i bhfolach ar long Padm\'{e}. Tachta\'{i}onn Anakin Padm\'{e} feargach go neamhfhiosach. Troideann Obi-Wan Anakin agus buann s\'{e}. Freastala\'{i}onn Obi-Wan d\'{a} chuid cosa agus f\'{a}gann s\'{e} \'{e} i mbruach abhainn l\'{a}ibhe ina dh\'{o}itear go m\'{o}r. Troideann Yoda an t-Impire Palpatine ar Coruscant go dt\'{i} go dtarla\'{i}onn an cath mar gheall air. Is Jedi cumhachtach \'{e} Yoda, ach n\'{i} f\'{e}idir leis an olc Impire Palpatine a chosc. T\'{e}ann Yoda ansin le Bail Organa agus t\'{e}ann Palpatine chuig Mustafar. \'{U}s\'{a}ideann an t-Impire Palpatine olc taobh dorcha na F\'{o}rsa le tuiscint go bhfuil Anakin i mbaol.

Casann Obi-Wan go Yoda chun athghr\'{u}th\'{u}. Tugann Padm\'{e} d\'{a} mhac agus d'in\'{i}on d\'{u}bailte a n-ainmn\'{i}onn s\'{i} Luke agus Leia. Braitheann Padm\'{e} br\'{o}n go gairid ina dhiaidh. Faigheann Palpatine Anakin d\'{o}ite go f\'{o}ill f\'{o}s beo ar Mustafar. Tar \'{e}is d\'{o} dul ar ais chuig Coruscant, d\'{e}ile\'{a}lfar le comhlacht m\'{a}inliachta Anakin agus cl\'{u}da\'{i}tear \'{e} in oireann arm\'{u}rtha dubh. B\'{i}onn Palpatine ag Anakin go mara\'{i}odh Padm\'{e} ina chlog. Is ceannm\'{a}istir olc \'{e} Palpatine agus f\'{a}gann moth\'{u} Anakin ar a ch\'{e}ile. Bhuaigh Palpatine; t\'{a} taobh dorcha na F\'{o}rsa anois ag Anakin. Idir an d\'{a} linn, oibr\'{i}onn Obi-Wan agus Yoda chun an c\'{u}pla a cheilt \'{o} thaobh dorcha na F\'{o}rsa, toisc gurb \'{e} an c\'{u}pla is d\'{o}chas ach amh\'{a}in le haghaidh saoirse. T\'{e}ann Yoda f\'{e}in chuig an bplain\'{e}ad Dagobah, agus maoir\'{i}onn Anakin agus an t-Impire Palpatine an d\'{e}ant\'{u}s an R\'{e}alt B\'{a}s. Uchta\'{i}onn Bail Organa Leia agus t\'{o}gann s\'{i} \'{i} chuig Alderaan. Taisteala\'{i}onn Obi-Wan le Luke go Tatooine. T\'{a} s\'{e} i gceist ag Obi-Wan f\'{e}achaint go l\'{a}idir ar Luke agus ar a theaghlach go dt\'{i} go mbeidh an t-am ceart d\'{u}shl\'{a}n a thabhairt don Impireacht.
}
\end{corpus}

\section{A More Complicated Example}\label{similarity}

We shall compute the similarity of meaning between ``Palpatine is a mastermind who turns Anakin to the dark side of the Force'' and its Irish equivalent, ``\textit{Is ceannm\'{a}istir a casann Anakin go taobh dorcha na F\'{o}rsa \'{e} Palpatine}''. The Irish sentence is assigned the following type:
\begin{gather*}
    \mbox{Is} \quad \mbox{ceannm\'{a}istir} \quad \mbox{a} \quad \mbox{casann} \quad \mbox{Anakin} \quad \mbox{go taobh dorcha na F\'{o}rsa} \quad \mbox{\'{e} Palpatine}\\
    s n^l n^l \quad n \quad n^r n n^{ll} s^l \quad s n^l n^l \quad n \quad n^r n  \quad n 
\end{gather*}
Abbreviating ``{taobh dorcha na F\'{o}rsa}'' as ``{TDNF}'', the reduction diagram is\footnote{Taking cues from the English ``who'' \cite{sadrz} regarding the depiction of ``{\it a}'' in the diagram.}:

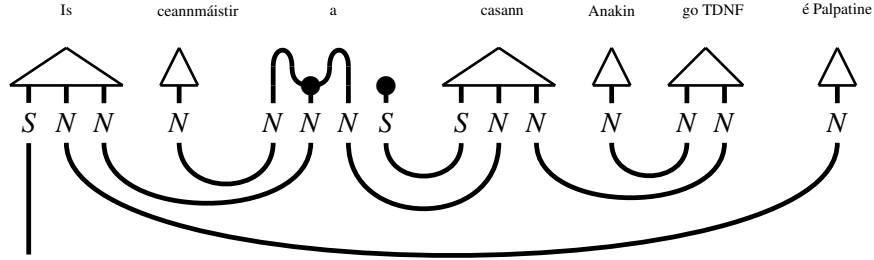
\begin{figure}[H] \centering
\begin{tikzpicture}[scale=0.5]
	\begin{pgfonlayer}{nodelayer}
		\node  (0) at (-6.5, 0.9999999) {$S$};
		\node  (1) at (-5.5, 0.9999999) {$N$};
		\node  (2) at (-4.5, 0.9999999) {$N$};
		
		\node  (3) at (15, 0.9999999) {$N$};
		
		\node  (4) at (-2.5, 0.9999999) {$N$};
		\node  (5) at (0, 1) {$N$};
		\node  (6) at (1, 1) {$N$};
		\node  (7) at (2, 1) {$N$};
		\node  (8) at (3, 1) {$S$};
		\node  (9) at (5, 1) {$S$};
		\node  (10) at (6, 1) {$N$};
		\node  (11) at (7, 1) {$N$};
		\node  (12) at (9, 1) {$N$};
		\node  (13) at (11, 1) {$N$};
		\node  (14) at (12, 1) {$N$};
		
		\node  (17) at (-6.5, 2) {};
		\node  (18) at (-5.5, 2) {};
		\node  (19) at (-4.5, 2) {};
		\node  (20) at (-5.5, 3) {};
		\node  (21) at (-7, 2) {};
		\node  (22) at (-4, 2) {};
		
		\node  (23) at (14.5, 2) {};
		\node  (24) at (15, 2) {};
		\node  (25) at (15.5, 2) {};
		\node  (26) at (15, 3) {};
		
		\node  (27) at (-3, 2) {};
		\node  (28) at (-2.5, 2) {};
		\node  (29) at (-2, 2) {};
		\node  (30) at (-2.5, 3) {};
		\node  (31) at (5, 2) {};
		\node  (32) at (6, 2) {};
		\node  (33) at (7, 2) {};
		\node  (34) at (4.5, 2) {};
		\node  (35) at (7.500001, 2) {};
		\node  (36) at (6, 3) {};
		\node  (37) at (8.499999, 2) {};
		\node  (38) at (9, 2) {};
		\node  (39) at (9.499999, 2) {};
		\node  (40) at (9, 3) {};
		\node  (41) at (10.5, 2) {};
		\node  (42) at (11, 2) {};
		\node  (43) at (12, 2) {};
		\node  (44) at (12.5, 2) {};
		\node  (45) at (11.5, 3) {};
		\node  (51) at (-6.5, -2.5) {};
		\node  (52) at (3, 2) {$\CIRCLE$};
		\node  (53) at (1, 2) {$\CIRCLE$};
		\node  (54) at (0, 2) {};
		\node  (55) at (2, 2) {};
		\node  (56) at (0.5000001, 2.5) {};
		\node  (57) at (1.5, 2.5) {};
		\node  (58) at (0, 2.5) {};
		\node  (59) at (2, 2.5) {};
		\node  (60) at (-5.5, 4) {\tiny  Is};
		
		\node  (61) at (15, 4) {\tiny  \'{e} Palpatine};
		
		\node  (62) at (-2, 4) {\tiny  ceannm\'{a}istir};
		\node  (63) at (1.75, 4) {\tiny  a\phantom{A}};
		\node  (64) at (6.25, 4) {\tiny  casann\phantom{A}};
		\node  (65) at (9, 4) {\tiny  Anakin};
		\node  (66) at (11.7, 3.97) {\tiny go TDNF};
	\end{pgfonlayer}
	\begin{pgfonlayer}{edgelayer}
		\draw [line width=1.8pt] (0) to (51.center);
		\draw [line width=1.8pt](17.center) to (0);
		\draw [line width=1.8pt](18.center) to (1);
		\draw [line width=1.8pt](19.center) to (2);
		\draw [line width=1pt](21.center) to (22.center);
		\draw [line width=1pt](22.center) to (20.center);
		\draw [line width=1pt](20.center) to (21.center);
		\draw [line width=1pt](23.center) to (25.center);
		\draw [line width=1pt](25.center) to (26.center);
		\draw [line width=1pt](26.center) to (23.center);
		\draw [line width=1.8pt](24.center) to (3);
		
		\draw [line width=1pt](27.center) to (28.center);
		\draw [line width=1pt](28.center) to (29.center);
		\draw [line width=1pt](27.center) to (30.center);
		\draw [line width=1pt](30.center) to (29.center);
		\draw [line width=1.8pt](28.center) to (4);
		\draw [bend right=90, looseness=1.50, line width=1.8pt] (4) to (5);
		\draw [bend right=90, looseness=1.50, line width=1.8pt] (8) to (9);
		\draw [line width=1.8pt](9) to (31.center);
		\draw [line width=1pt](34.center) to (35.center);
		\draw [line width=1pt](35.center) to (36.center);
		\draw [line width=1pt](36.center) to (34.center);
		\draw [line width=1.8pt](32.center) to (10);
		\draw [line width=1.8pt](33.center) to (11);
		\draw [line width=1pt](37.center) to (39.center);
		\draw [line width=1pt](39.center) to (40.center);
		\draw [line width=1pt](40.center) to (37.center);
		\draw [line width=1.8pt](38.center) to (12);
		\draw [line width=1pt](41.center) to (44.center);
		\draw [line width=1pt](44.center) to (45.center);
		\draw [line width=1pt](45.center) to (41.center);
		\draw [line width=1.8pt](42.center) to (13);
		\draw [line width=1.8pt](43.center) to (14);
		\draw [bend right=90, looseness=1.50, line width=1.8pt] (12) to (13);
		\draw [bend right=90, looseness=1.00, line width=1.8pt] (2) to (6);
		
		\draw [bend right=90, looseness=0.5, line width=1.8pt] (1) to (3);
		
		\draw [bend right=90, looseness=1.50, line width=1.8pt] (7) to (10);
		\draw [bend right=90, looseness=1.00, line width=1.8pt] (11) to (14);
		\draw [line width=1.8pt](52.center) to (8);
		\draw [line width=1.8pt](53.center) to (6);
		\draw [line width=1.8pt](54.center) to (5);
		\draw [line width=1.8pt](55.center) to (7);
		\draw [bend right=45, looseness=1.25, line width=1.8pt] (56.center) to (53.center);
		\draw [bend right=45, looseness=1.25, line width=1.8pt] (53.center) to (57.center);
		\draw [line width=1.8pt](58.center) to (54.center);
		\draw [line width=1.8pt](59.center) to (55.center);
		\draw [bend left=90, looseness=3.00, line width=1.8pt] (58.center) to (56.center);
		\draw [bend left=90, looseness=3.00, line width=1.8pt] (57.center) to (59.center);
	\end{pgfonlayer}
\end{tikzpicture}
\end{figure}
\vspace{-7mm}
and is simplified to:

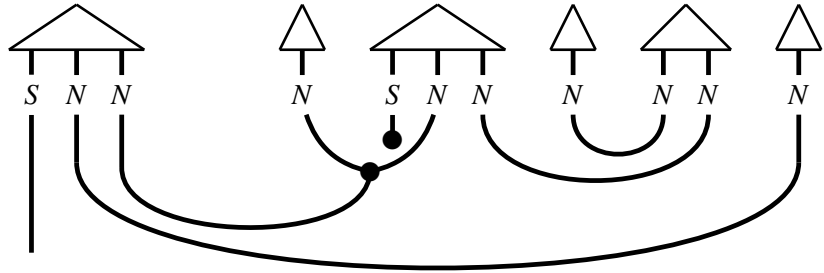
\begin{figure}[H] \centering
\begin{tikzpicture}[scale=0.6]
	\begin{pgfonlayer}{nodelayer}
		\node  (0) at (-10, 2) {$S$};
		\node  (1) at (-9, 2) {$N$};
		\node  (2) at (-8, 2) {$N$};
		\node  (3) at (-4, 2) {$N$};
		\node  (4) at (-2, 2) {$S$};
		\node  (50) at (-2, 1) {$\CIRCLE$};
		\node  (5) at (-1, 2) {$N$};
		\node  (6) at (0, 2) {$N$};
		\node  (7) at (2, 2) {$N$};
		\node  (8) at (4, 2) {$N$};
		\node  (9) at (5, 2) {$N$};
		\node  (12) at (-10, 3) {};
		\node  (13) at (-9, 3) {};
		\node  (14) at (-8, 3) {};
		\node  (15) at (-10.5, 3) {};
		\node  (16) at (-7.5, 3) {};
		
		\node  (17) at (-9, 4) {};
		
		\node  (18) at (-4.5, 3) {};
		\node  (19) at (-4, 3) {};
		\node  (20) at (-3.5, 3) {};
		\node  (21) at (-4, 4) {};
		\node  (22) at (-2.5, 3) {};
		\node  (23) at (-2, 3) {};
		\node  (24) at (-1, 3) {};
		\node  (25) at (0, 3) {};
		\node  (26) at (0.5, 3) {};
		\node  (27) at (-1, 4) {};
		\node  (28) at (1.5, 3) {};
		\node  (29) at (2, 3) {};
		\node  (30) at (2.5, 3) {};
		\node  (31) at (2, 4) {};
		\node  (32) at (3.5, 3) {};
		\node  (33) at (4, 3) {};
		\node  (34) at (5, 3) {};
		\node  (35) at (5.5, 3) {};
		\node  (36) at (4.5, 4) {};
		\node  (42) at (-10, -1.5) {};
		
		\node  (43) at (7, 2) {$N$};
		
		\node  (44) at (7, 3) {};
		\node  (45) at (7.5, 3) {};
		\node  (46) at (7, 4) {};
		\node  (47) at (6.5, 3) {};
		
		\node  (48) at (-2.5, 0.3) {$\CIRCLE$};
		\node  (49) at (-8, 0.4) {};
		
		\node  (52) at (-9, 0.5) {};
		\node  (51) at (7, 0.5) {};
	\end{pgfonlayer}
	\begin{pgfonlayer}{edgelayer}
		\draw [line width=1pt](15.center) to (16.center);
		\draw [line width=1pt](16.center) to (17.center);
		\draw [line width=1pt](17.center) to (15.center);
		\draw [line width=1pt](18.center) to (20.center);
		\draw [line width=1pt](20.center) to (21.center);
		\draw [line width=1pt](21.center) to (18.center);
		\draw [line width=1pt](22.center) to (26.center);
		\draw [line width=1pt](26.center) to (27.center);
		\draw [line width=1pt](27.center) to (22.center);
		\draw [line width=1pt](28.center) to (30.center);
		\draw [line width=1pt](30.center) to (31.center);
		\draw [line width=1pt](31.center) to (28.center);
		\draw [line width=1pt](32.center) to (35.center);
		\draw [line width=1pt](35.center) to (36.center);
		\draw [line width=1pt](36.center) to (32.center);
		\draw [line width=1.8pt](12.center) to (0);
		\draw [line width=1.8pt](0) to (42.center);
		\draw [line width=1.8pt](13.center) to (1);
		\draw [line width=1.8pt](14.center) to (2);
		\draw [line width=1.8pt](19.center) to (3);
		\draw [line width=1.8pt](23.center) to (4);
		\draw [line width=1.8pt](24.center) to (5);
		\draw [line width=1.8pt](25.center) to (6);
		\draw [line width=1.8pt](29.center) to (7);
		\draw [line width=1.8pt](33.center) to (8);
		\draw [line width=1.8pt](34.center) to (9);
		
		\draw [line width=1.8pt](1) to (52.center);
		\draw [line width=1.8pt](43) to (51.center);
		
		\draw [bend right=90, looseness=1.50, line width = 1.8pt] (7) to (8);
		\draw [line width=1pt](47.center) to (45.center);
		\draw [line width=1pt](45.center) to (46.center);
		\draw [line width=1pt](46.center) to (47.center);
		\draw [line width=1.8pt](44.center) to (43);
		\draw [bend right=90, looseness=0.5, line width = 1.8pt] (52.center) to (51.center);
		\draw [bend right=90, looseness=1.00, line width = 1.8pt] (6) to (9);
		\draw [line width=1.8pt](2) to (49.center);
		\draw [line width=1.8pt](4) to (50.center);
		\draw [bend right=90, looseness=0.8, line width = 1.8pt] (49.center) to (48.center);
		\draw [bend left, looseness=1.00, line width = 1.8pt] (48.center) to (3);
		\draw [bend right, looseness=1.00, line width = 1.8pt] (48.center) to (5);
	\end{pgfonlayer}
\end{tikzpicture}
\end{figure}
\vspace{-7mm}

This diagram corresponds to the map
\begin{align*}
f = (1_S \otimes \epsilon_N) &\circ (1_S \otimes 1_N \otimes \epsilon_N \otimes 1_N) \circ (1_S \otimes 1_N \otimes 1_N \otimes \mu_N \otimes \epsilon_N \otimes 1_N) \\
&\circ (1_S \otimes 1_N \otimes 1_N \otimes 1_N \otimes i_S \otimes 1_N \otimes 1_N \otimes \epsilon_N \otimes 1_N \otimes 1_N).    
\end{align*}

Note in this sentence the verb \emph{casann} (``turns'') has been modified to follow the rule \emph{Verb Object Subject}. Therefore we will use the transpose of the matrix $C^{\mbox{\tiny cas}}$ to accommodate this change. The meaning vector of the sentence ``\textit{Is ceannma\'{a}istir a cassan Anakin go taobh dorcha na F\'{o}rsa \'{e} Palpatine}'' is:
\begin{align*}
&\overrightarrow{\mbox{{\it Is ceannm\'{a}istir a casann Anakin go taobh dorcha na F\'{o}rsa \'{e} Palpatine}}}\\
&= 330 \overrightarrow{n}_2 \otimes \overrightarrow{n}_1 + 40 \overrightarrow{n}_2 \otimes \overrightarrow{n}_2.
\end{align*}
The author has excluded the calculations of the matrices $(C^{\mbox{\tiny cas}})^T$ and $C^{\mbox{\tiny go TDNF}}$ for brevity, but these are calculated as per \S\ref{irishvect}. Taking the inner product,
\begin{align*}
&\langle \overrightarrow{\mbox{Palpatine is a mastermind who turns Anakin to the dark side of the Force}} \mbox{ } | \\
&\qquad\overrightarrow{\mbox{{\it Is ceannm\'{a}istir a casann Anakin go taobh dorcha na F\'{o}rsa \'{e} Palpatine}}} \rangle = 106880.
\end{align*}
The length of the former is 103424, and the length of the latter is 110500. Therefore their similarity score is $\tfrac{106880}{\sqrt{103424\cdot 110500}} = 0.999$; very high. \\

Suppose we thought the translation of ``\textit{Is ceannm\'{a}istir a casann Anakin go taobh dorcha na F\'{o}rsa \'{e} Palpatine}'' was ``\textbf{The Emperor} is a mastermind who turns Anakin to the dark side of the Force''. The calculations tell us:
\begin{align*}
&\langle \overrightarrow{\mbox{The Emperor is a mastermind who turns Anakin to the dark side of the Force}} \mbox{ } | \\
&\qquad\overrightarrow{\mbox{{\it Is ceannm\'{a}istir a casann Anakin go taobh dorcha na F\'{o}rsa \'{e} Palpatine}}} \rangle = 534400.
\end{align*}

The length of the former is 2593792, and the length of the latter is 110500. Therefore the similarity score of the sentences is $\tfrac{534400}{\sqrt{2593792\cdot 110500}} = 0.998$; nearly as high as the actual translation. Of course, as Palpatine \emph{is} the Emperor, one could argue this is still a valid translation.

\section{Corpora for a Conceptual Space Model of Meaning}\label{spaceappendix}
We shall make the same assumptions regarding this corpus as in \emph{Appendix \ref{sweng}}.

\begin{corpus}\label{spacecorp}
Venus is a planet in the solar system. Venus has a solid and rocky surface. Venus is called Earth's sister because it is nearly the same size as Earth. Venus is very hot and the pressure on its surface is high. Venus is bright in the night sky and looks like a ball.

Jupiter, another planet in the solar system, also looks like a ball. Jupiter sits in outer space. The size of Jupiter is very large; it is the largest planet in the solar system. Jupiter is large and gassy. Jupiter is primarily orange and brown and red in colour. Jupiter is far away from Earth. It is very windy on Jupiter and also freezing cold. Jupiter is very bright in the night sky.

Mars is a planet next to Earth. Mars is coloured very red, and brown and orange. Mars is cold, but not very cold. Mars is smaller than Earth. Mars is rocky like Venus and Earth. Mars sits in outer space.

Apples are fruits. Apples are round and soft. Apples can be red or green, and some apples taste bitter and other apples taste sweet. An apple looks like a ball.

The Sun is a star, not a planet. It sits in the centre of the solar system. The Sun is the brightest thing in the sky. The Sun is huge and very hot. The Sun is round and also looks like a ball. The gravity on the Sun is very strong, meaning it is very dense.
\end{corpus}

\noindent Here is the same corpus, but in Irish:

\begin{corpus}\label{spacecorpirish}{
Is \'{i} V\'{e}ineas pl\'{a}n\'{e}ad sa ghrianch\'{o}ras. T\'{a} dromchla tathagach agus carraigeach ag V\'{e}ineas. Glaotar V\'{e}ineas deirfi\'{u}r an Domhan \'{i} mar t\'{a} s\'{i} beagnach an m\'{e}id c\'{e}anna leis an Domhan. T\'{a} s\'{e} an-te ar V\'{e}ineas agus t\'{a} an br\'{u} ar a dromchla ard. T\'{a} V\'{e}ineas geal i sp\'{e}ir na ho\'{i}che agus breathna\'{i}onn s\'{i} cos\'{u}il le liathr\'{o}id.

Breathna\'{i}onn I\'{u}patar, pl\'{a}n\'{e}ad eile sa ghrianch\'{o}ras, cos\'{u}il le liathr\'{o}id freisin. Su\'{i}onn I\'{u}patar i sp\'{a}s seachtrach. T\'{a} I\'{u}patar an-mh\'{o}r; t\'{a} s\'{e} an pl\'{a}n\'{e}ad is m\'{o} sa ghrianch\'{o}ras. T\'{a} I\'{u}patar m\'{o}r agus d\'{e}anta s\'{e} as g\'{a}is. T\'{a} I\'{u}patar go pr\'{i}omha or\'{a}iste agus donn agus dearg i ndath. T\'{a} I\'{u}patar i bhfad gc\'{e}in \'{o} an Domhan. T\'{a} s\'{e} an-ghaothmhar ar I\'{u}patar agus an-fhuar freisin. T\'{a} I\'{u}patar an-gheal i sp\'{e}ir na ho\'{i}che.

Is \'{e} Mars pl\'{a}in\'{e}id in aice leis an Domhan. T\'{a} Mars daite an-dearg, agus donn agus or\'{a}iste. T\'{a} s\'{e} fuar ar Mars, ach n\'{i}l s\'{e} an-fhuar. T\'{a} Mars n\'{i}os l\'{u} n\'{a} an Domhan. T\'{a} Mars carraigeach cos\'{u}il le V\'{e}ineas agus an Domhan. Su\'{i}onn Mars i sp\'{a}s seachtrach.

Is tortha\'{i} iad \'{u}lla. T\'{a} \'{u}lla liathr\'{o}ideach agus bog. F\'{e}adfaidh \'{u}lla a bheith dearg n\'{o} glas, agus t\'{a} blas searbh ar roinnt \'{u}ill agus blas milis ar \'{u}lla eile. Breathna\'{i}onn \'{u}ll cos\'{u}il le liathr\'{o}id.

Is r\'{e}alta \'{i} an Grian, n\'{i} phl\'{a}in\'{e}id. T\'{a} s\'{i} suite i l\'{a}r an ch\'{o}rais ghr\'{e}ine. Is \'{i} an grian an rud is gile sa sp\'{e}ir. T\'{a} an grian ollmh\'{o}r agus an-te. T\'{a} an grian liathr\'{o}ideach agus breathna\'{i}onn s\'{i} ar liathr\'{o}id fresin. T\'{a} an imtharraingt ar an ghrian an-l\'{a}idir, rud a chialla\'{i}onn go bhfuil s\'{i} an-dl\'{u}th.}
\end{corpus}

\section{Additional assignment of noun spaces arising from \textit{Corpus \ref{spacecorpirish}}.}\label{rename}

Organising the information of \emph{Corpus \ref{spacecorpirish}} into a table we obtain:
\vspace{-3mm}
\begin{center}
\begin{longtable}{ |>{\em}c|c|>{\em}p{5cm}|p{5cm}| } 
 \hline
 {V\'{e}ineas} & \textbf{Adjectives} & {tathagach}, {carraigeach}, {beagnach an m\'{e}id c\'{e}anna leis an Domhan}, {an-te}, {br\'{u} \dots ard}, {geal}. & solid, rocky, same size as Earth, hot, high pressure, bright.\\
 \hline
 & \textbf{Nouns} & {pl\'{a}n\'{e}ad}, { deirfi\'{u}r an Domhan}, {liathr\'{o}id}. & planet, Earth's sister, ball.\\
 \hline
  {I\'{u}patar} & \textbf{Adjectives} & {an-mh\'{o}r}, {d\'{e}anta as g\'{a}is}, {or\'{a}iste}, {donn}, {dearg}, {i bhfad i gc\'{e}in}, { an-ghaothmhar}, {an-fhuar}, {an-gheal}. & very large, gassy, orange, brown, red, far away, windy, freezing, very bright.\\
 \hline
 & \textbf{Nouns} & {pl\'{a}n\'{e}ad}, {sp\'{a}s seachtrach}, {liathr\'{o}id}. & planet, outer space, ball.\\
 \hline
  {Mars} & \textbf{Adjectives} & {an-dearg}, {or\'{a}iste}, {donn}, {fuar}, {n\'{i}os l\'{u} n\'{a} an Domhan}, {carraigeach}. & very red, brown, orange, cold, smaller than Earth, rocky.\\
 \hline
 & \textbf{Nouns} & {pl\'{a}n\'{e}ad}, {sp\'{a}s seachtrach}. & planet, outer space.\\
 \hline
  {\'{U}ll} & \textbf{Adjectives} & {liathr\'{o}ideach}, {bog}, {dearg}, {gl\'{a}s}, {searbh}, {milis}. & round, soft, red, green, bitter, sweet.\\
 \hline
 & \textbf{Nouns} & {tortha\'{i}}, {liathr\'{o}id}. & fruit, ball.\\
 \hline
  {Grian} & \textbf{Adjectives} & {an rud is gile}, {ollmh\'{o}r}, {an-te}, {liathr\'{o}ideach}, {an-dl\'{u}th}. & brightest, huge, very hot, round, very dense. \\
 \hline
 & \textbf{Nouns} & {r\'{e}alta}, {liathr\'{o}id}. & star, ball.\\
 \hline
\end{longtable}
\end{center}

\vspace{-7mm}
\noindent Recall that, if an adjective class is not mentioned, its corresponding noun space is set to $[0,1]$.

{\small
\hspace{-4mm}\begin{minipage}{0.45\textwidth}
\begin{align*}
\mbox{(2) }&\mbox{\it \textbf{{I\'{u}patar}.}}\\
&N_{\mbox{\tiny  dimension}} = Conv(\mbox{\it {an-mh\'{o}r}}) = \{0.8\},\\
&N_{\mbox{\tiny  colour}} = Conv(\mbox{\it {or\'{a}iste}} \cup \mbox{\it {donn}} \cup \mbox{\it {dearg}})\footnotemark,\\
&N_{\mbox{\tiny  intensity}} = Conv(\mbox{\it {an-geal}}) = \{0.7\},\\
&N_{\mbox{\tiny  temperature}} = Conv(\mbox{\it {an-fuar}}) = \{0.1\},\\
&N_{\mbox{\tiny  density}} = Conv(\mbox{\it {d\'{e}anta as g\'{a}is}}) = \{0.1\}.
&\mbox{ }\\
&\mbox{ }\\
\end{align*}
\end{minipage}\hspace{4mm}\begin{minipage}{0.45\textwidth}
\begin{align*}
\mbox{(3) }&\mbox{\it \textbf{Mars.}}\\
&N_{\mbox{\tiny  dimension}} = Conv(\mbox{\it {n\'{i}os l\'{u} n\'{a} an Domhan}}) = \{0.25\},\\
&N_{\mbox{\tiny  colour}} = Conv(\mbox{\it {dearg}} \cup \mbox{\it {donn}} \cup \mbox{\it {or\'{a}iste}}),\\
&N_{\mbox{\tiny  temperature}} = Conv(\mbox{\it {fuar}}) = \{0.4\},\\
&N_{\mbox{\tiny  texture}} = Conv(\mbox{\it {carraigeach}}) = \{0.9\}.
&\mbox{ }\\
&\mbox{ }\\
&\mbox{ }\\
\end{align*}
\end{minipage}}\footnotetext{The RGB values we use for \emph{or\'{a}iste}, \emph{donn} and \emph{dearg} are (255,165,0), (153,76,0), and (255,0,0) respectively.}

\vspace{-10mm}
{\small
\hspace{-13mm}\begin{minipage}{0.45\textwidth}
\vspace{-5mm}\begin{align*}
\mbox{(4) }&\mbox{\it \textbf{\'{U}ll.}}\\
&N_{\mbox{\tiny  colour}} = Conv(\mbox{\it {dearg}} \cup \mbox{\it {gl\'{a}s}}),\\
&N_{\mbox{\tiny  taste}} = Conv(\mbox{\it {searbh}} \cup \mbox{{milis}}),\\
&N_{\mbox{\tiny  texture}} = Conv(\mbox{\it {bog}}) = \{0.4\}.
\end{align*}
\end{minipage}\hspace{6mm}\begin{minipage}{0.45\textwidth}
\begin{align*}
\mbox{(5) }&\mbox{\it \textbf{Grian.}}\\
&N_{\mbox{\tiny  dimension}} = Conv(\mbox{\it {ollmh\'{o}r}}) = \{0.9\},\\
&N_{\mbox{\tiny  intensity}} = Conv(\mbox{\it {an rud is gile}}) = \{1\},\\
&N_{\mbox{\tiny  temperature}} = Conv(\mbox{\it {an-te}}) = \{0.85\},\\
&N_{\mbox{\tiny  density}} = Conv(\mbox{\it {an-dl\'{u}th}}) = \{1\}.
\end{align*}
\end{minipage}}

\vspace{4mm}
For $K = 1, \dots, 5$, $D^K_{\mbox{\tiny  adj}}$ is the tensor product of the $N_{\mbox{\scriptsize(--)}}$ spaces of item (K). These values were assigned according to the author's own preference, however they can be assigned different values according to each readers' wishes. 
\end{appendices}

\end{document}